\documentclass[runningheads]{llncs}
\usepackage{epsfig}
\usepackage{graphicx}
\usepackage{amsmath}
\usepackage{amssymb}
\usepackage{subcaption}

\usepackage{lmodern}

\usepackage[utf8]{inputenc} 
\usepackage[T1]{fontenc}    
\usepackage{url}            
\usepackage{booktabs}       
\usepackage{amsfonts}       
\usepackage{nicefrac}       
\usepackage{microtype}      

\usepackage{caption}
\usepackage{multirow}
\usepackage{xspace}
\usepackage{mathtools}
\usepackage{wasysym}

\usepackage{marvosym}

\usepackage[table,xcdraw]{xcolor}
\usepackage{pdfcomment}
\usepackage{lipsum}
\usepackage{algorithm,algpseudocode}
\usepackage{wrapfig}

\usepackage{hyperref}
\hypersetup{breaklinks=true,colorlinks}

\def\ECCVSubNumber{1851}  

\def\ie{{i.e.}}
\def\cp{{\mathbf{cp}}}

\makeatother

\newcommand{\SF}[1]{{\color{magenta}{[Sanja: #1]}}}

\newcommand{\Xiaohui}[1]{{\color{violet}{[Xiaohui: #1]}}}
\newcommand{\ours}{ScribbleBox}

\begin{document}
\title{ScribbleBox: Interactive Annotation Framework for Video Object Segmentation}
\titlerunning{ScribbleBox} 
\authorrunning{B. Chen, H. Ling, et al.} 
\author{Bowen Chen$^{1}$\thanks{authors contributed equally}, Huan Ling$^{1,2,3, \star}$, Xiaohui Zeng$^{1,2}$, \\Jun Gao$^{1,2,3}$, Ziyue Xu$^{1}$, Sanja Fidler$^{1,2,3}$ }

\institute{$^1$University of Toronto \hspace{1.5mm} $^2$Vector Institute \hspace{1.5mm}   $^3$NVIDIA \\
\email{\{chenbowen, linghuan, xiaohui, jungao, fidler\}@cs.toronto.edu}
\\
\email{\{ziyue.xu\}@mail.utoronto.ca}}


\maketitle
\begin{abstract}
Manually labeling video datasets for segmentation tasks is extremely time consuming. We introduce {\ours}, an interactive framework for annotating object instances with masks in videos with a significant boost in efficiency. In particular, we split annotation into two steps: annotating objects with tracked boxes, and labeling masks inside these tracks. We introduce automation and interaction in both steps. Box tracks are annotated efficiently by approximating the trajectory using a parametric curve with a small number of control points which the annotator can interactively correct. Our approach tolerates a modest amount of noise in  box placements, thus typically requiring only a few clicks to annotate a track to a sufficient accuracy.  Segmentation masks are corrected via scribbles which are propagated through time. We show significant performance gains in annotation efficiency over past work. We show that our {\ours} approach reaches 88.92\% J\&F on DAVIS2017 with an average of 9.14 clicks per box track, and only 4 frames requiring scribble annotation in a video of 65.3 frames on average.    
\end{abstract}

\section{Introduction}
\label{sec:intro}

Video is one of the most common forms of visual media. It is used to entertain us (film, tv series), inform us (news), educate us (video lectures), connect us (video conferencing), and attract our interest via TV commercials and social media posts.
Video is also a crucial modality for robotic applications such as self-driving cars, security applications, and patient monitoring in healthcare. 

One of the fundamental tasks common to a variety of applications is the ability to segment and track individual objects across the duration of the video. 
However, the success of existing methods in this task is limited due to the fact that training data is hard to obtain. Labeling only a single object in a single frame can take up to a minute~\cite{polyrnn,ChenCVPR14}, thus annotating the full video is prohibitively time consuming. 
This fact also presents a major roadblock for video content editing where end-users may want to segment a particular object/person and replace the background with an alternative. Most film studios thus still mainly resort to the use of ``green-screen'' during recording. 
Our interest here is in interactive annotation to assist human users in segmenting objects in videos.

Prior work on human assisted interactive annotation has addressed both the image~\cite{Ling_2019_CVPR,polyrnnpp,wang2019delse,dextr,Rother2004SIGGRAPH,Mortensen} and video domains~\cite{chen2018blazingly,Benard2018InteractiveVO,mahadevan2018iteratively,Levinkov2016InteractiveMV,Caelles_arXiv_2018,oh2019fast}. The prevailent image-based approaches employ grabCut-like techniques using scribbles as human feedback~\cite{Rother2004SIGGRAPH}, track object boundaries via intelligent scissors~\cite{Mortensen}, interactively edit predicted vertices of a polygon outlining the object~\cite{Ling_2019_CVPR,polyrnnpp}, and providing human clicks on the erroneously predicted object boundary as a heatmap to  a DeepLab-like architecture~\cite{dextr,wang2019delse}. In video, methods either use online fine-tuning~\cite{DAVIS2018-Interactive-2nd,DAVIS2019-Interactive-3rd} of the model given a new user-annotated frame, or propose ways to  propagate scribbles provided by the user in one frame to other frames. However, in current scribble propagation approaches a human-provided scribble typically only affects neighboring frames, and oftentimes the scribbles are ignored entirely by the neural network that processes them. 


\begin{figure}[t!]
\begin{minipage}{0.501\linewidth}
\begin{center}
\includegraphics[width=1\linewidth,trim=0 0 0 1,clip]{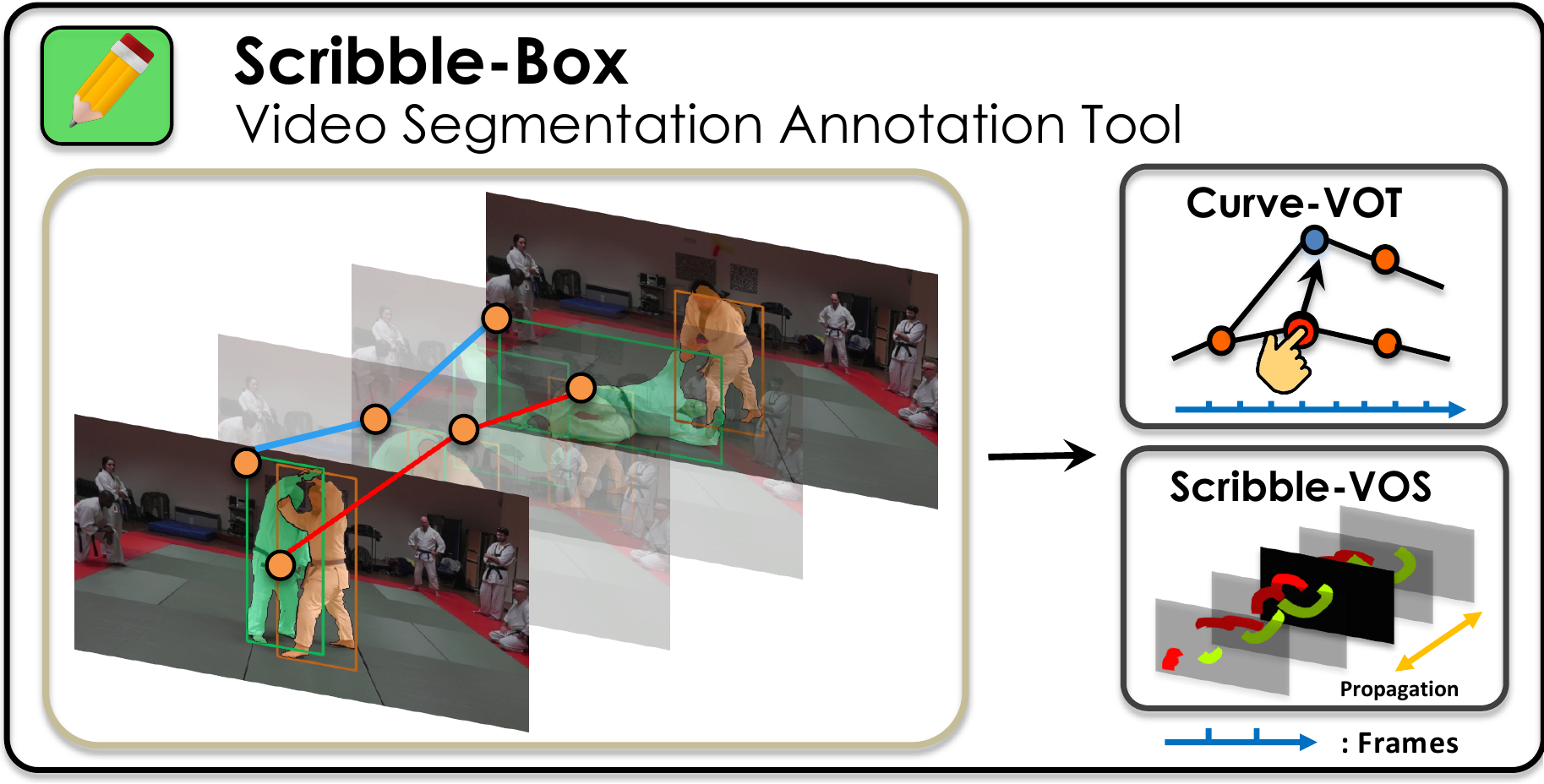} 
\end{center}
\end{minipage}
\hspace{1mm}
\begin{minipage}{0.46\linewidth}
\caption{\footnotesize {\bf ScribbleBox}: Interactive framework for annotating object masks in videos. Our approach splits annotation into two steps: interactive box tracking where the user corrects keyframe boxes (Curve-VOT), and interactive segmentation inside tracked boxes with human-provided scribbles (Scribble-VOS).} 
\label{fig:intro}
\end{minipage}
\end{figure}

In this paper, we introduce {\ours}, a novel approach to interactive annotation of object instances in videos. In particular, we propose to  split the task of segmenting objects into two simpler tasks:  annotating and tracking a loose box around each object across frames, and segmenting the object in each tracked box. We make both steps interactive. For tracking, we represent object's motion using a sequence of linear motions. Our key contribution is to optimize for the control points defining these motions, and allow the user to interactively correct the erroneous control points. We perform local adjustments to allow for the boxes to slightly deviate from the piece-wise linearly interpolated track. 
To segment an object inside each tracked box, we exploit scribbles as a form of human input. We present a new scribble propagation network and a simulation scheme that encourages the network to accurately propagate human input through the video. In a user study, we show that our approach achieves about 4\% higher J\&F score compared to the current state-of-the-art interactive baseline IPN~\cite{oh2019fast} given the same amount of annotation time.

 Online demo and tool will be released at \href{http://www.cs.toronto.edu/~linghuan/scribblebox/}{Project Page}

\section{Related Work}
\label{sec:related}

Literature on object tracking/segmentation is vast. Since automatic tracking and segmentation is not our contribution we limit our review to interactive methods. 


\textbf{Interactive Visual Object Tracking (VOT):}
In the traditional tracking annotation protocol, annotators are asked to determine keyframes and label the object's box in these frames.  Other frames are then automatically annotated via either linear interpolation~\cite{5459289}, or performing a shortest-path interpolation between manual annotations~\cite{CrowdsourcedVideo}. The annotator typically needs to go back and forth in determining the keyframes such that the final full track is accurate.  
To introduce intelligence in video annotation protocol and reduce cost,~\cite{ActiveVideo} exploits active learning to choose which frames to annotate. More recent and related to our method, PathTrack~\cite{PathTrack} presents an efficient framework to annotate tracking by tracing each object with a mouse cursor as the video plays. 
However, there is an ambiguity in determining the scale of the object which they try to automatically account for. In our work, the annotator does not need to watch the full video, and is only asked to inspect the automatically determined keyframes. 
While playing the in between segments in fast-forward mode is desired for verification, we experimentally show that no further inspection is typically needed, as our full method deals with a substantial amount of noise. 


\textbf{Interactive Video Object Segmentation (VOS):} 
With the recently introduced datasets~\cite{Caelles_arXiv_2018}, there has been an increased interest in designing both automatic and interactive methods for video object segmentation. An interactive method expects user input, either clicks or scribbles, to refine the predicted output masks. For efficiency, human feedback is propagated to nearby frames. 

Several earlier approaches  employed graph cut techniques~\cite{DBLP:journals/tog/WangBCAC05,DBLP:conf/iccv/PriceMC09,DBLP:journals/tog/LiSS05,DBLP:journals/tog/BaiWSS09}. In~\cite{DBLP:journals/tog/WangBCAC05}, user's input spans  three dimensions. Video is treated as a spatiotemporal graph and a preprocessing step is required to reduce the number of nodes for the min-cut problem. A much faster method is LIVEcut~\cite{DBLP:conf/iccv/PriceMC09} where the user selects and corrects frames which are propagated forward frame by frame. 
In~\cite{DBLP:conf/iccv/BaiS07}, an image with foreground and background scribble annotations is converted into a weighted graph, where the weights between nodes (pixels) are computed based on features like color or location. A pixel is classified based on its shortest geodesic distance to scribbles. 
JFS~\cite{DBLP:conf/iccv/NagarajaSB15} propagates user annotations through the computed point trajectory and classifies the remaining pixels with a random walk algorithm. 
Modern methods~\cite{chen2018blazingly,dextr,Benard2018InteractiveVO,Caelles_arXiv_2018,DAVIS2018-Interactive-2nd,oh2019fast} employ neural networks for interactive VOS. 
~\cite{chen2018blazingly} formulates the segmentation task as a pixel-wise retrieval problem and supports different kinds of user input such as masks, clicks and scribbles. 
\cite{Caelles_arXiv_2018} proposes an Interactive Segmentation Track for the DAVIS2018 Challenge along with baselines. The first baseline is based on an online-learning VOS model OSVOS~\cite{DBLP:conf/cvpr/CaellesMPLCG17}, while the second baseline trains a SVM classifier on pixels that are annotated with scribbles. Top performers in the challenge include SiamDLT~\cite{DAVIS2018-Interactive-2nd} which performs similarity learning together with online fine-tuning on the user-provided scribbles and dense CRF as post-processing.

The winner of the DAVIS2018 Challenge, IPN~\cite{oh2019fast} proposes an interactive framework by employing scribbles to correct masks and propagate corrections to the neighboring frames implicitly. 
Different from IPN, we incorporate tracking in the annotation process, and model scribble propagation explicitly: we directly predict scribbles for other frames based on the user's input, and employ a  training strategy that encourages our network to utilize scribbles effectively. Note that our mask propagation module shares similarity with the recent mask propagation model Space-Time Memory Networks~\cite{oh2019video}. We employ Graph Convolutional Networks while they adopt a memory-network architecture. 
\section{Our Approach}
\label{sec:method}


\begin{figure}[t!]
\begin{minipage}{0.59\linewidth}
\begin{center}
\includegraphics[width=1.1\linewidth,trim=0 0 0 0,clip]{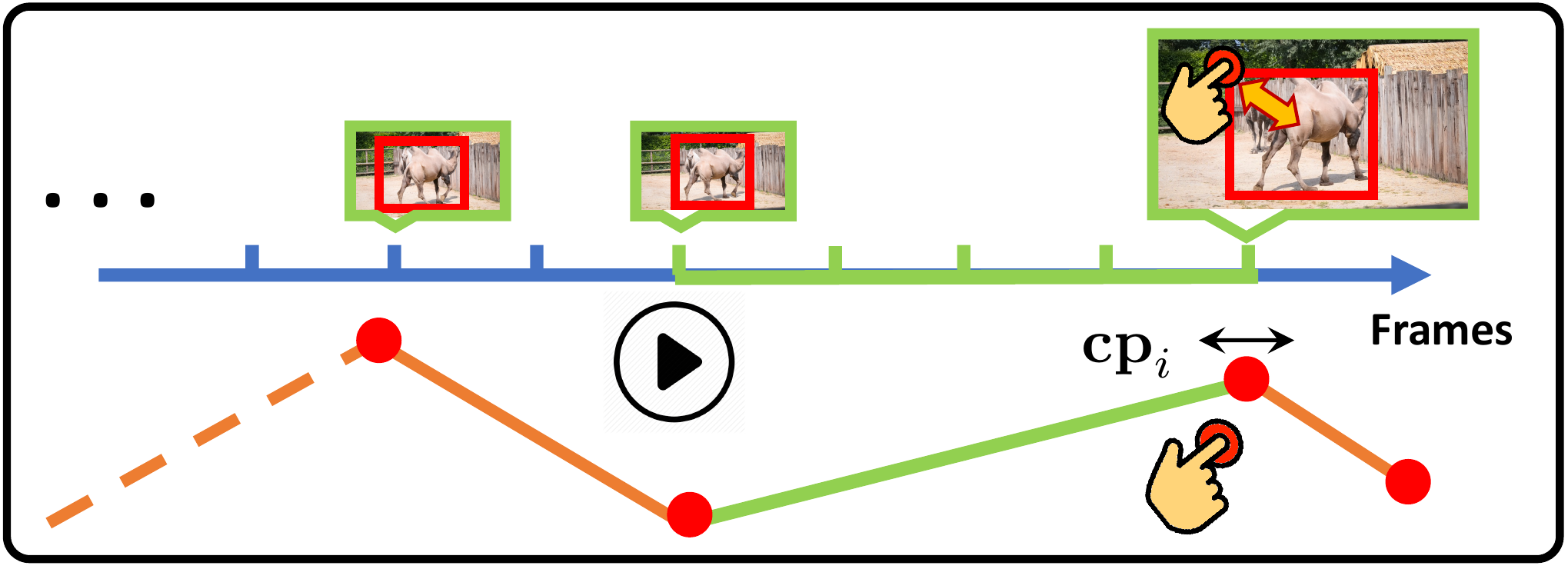}
\end{center} 
\end{minipage}
\hfill
\begin{minipage}{0.34\linewidth}
\caption{\footnotesize {\bf Curve-VOT:} Annotators inspect automatically determined keyframes and adjust control points of a parametric curve. They can fast-forward the video in between keyframes to ensure the object is adequately tracked.} 
\label{fig:interface}
\end{minipage}
\end{figure}

The goal of our work is to efficiently label object tracks in videos with accurate masks. 
We split annotation into two steps: 1) box tracking, and 2) mask labeling given the tracked boxes. We argue that it is easier to segment the object inside cropped regions rather than the full image both for the human labeler as well as automatic methods as they can more explicitly exploit shape priors and correlations of motion across time. 

We introduce automation and human-in-the-loop  interaction in both stages, aiming to achieve the highest level of labeling efficiency. We represent the box track with a parametric curve (polygon or spline~\cite{gao2019deepspline,Ling_2019_CVPR}) exploiting the fact that many motions are roughly linear locally. The curve is represented with a small set of control points which are obtained automatically. The annotator is then only shown the frames closest to the control points, and in case of errors a  control point can be adjusted. We exploit this information to re-estimate the control points in the subsequent parts of the video. This process can be done iteratively.  
Importantly, the annotated tracked boxes obtained from this step can be quite loose, and mainly serve to roughly delineate where the object is in the frames.  Our second step, the mask labeling step, is able to tolerate such noise and produces tracked masks with very few interactions with the annotator. 

In the second annotation step, we first automatically label the object's masks.
The annotator then inspects the video of the segmented object 
and provides feedback for the frames where errors occur by drawing scribbles in the erroneous areas. We automatically re-predict the frame and propagate this information to the subsequent frames. We explain both steps in more detail next. 


\subsection{Interactive Tracking Annotation}
\label{sec:VOT}
We here introduce \emph{Curve-VOT}, our interactive approach for tracking annotation which tries to reduce the number of clicks required to annotate boxes around an object of interest in a video. We approximate the trajectory of the box using a parametric curve which is estimated interactively. Specifically, we approximate a $M$-frame trajectory $J$ as a polygonal curve $P$ with a fixed number of control points, $N$. We obtain this curve automatically, and allow the annotator to edit the control points by moving them in space and time, and add or remove points.

Note that our approximation assumes that the motion of the box between two control points is linear. This assumption typically holds locally, but may be violated if the temporal gap between the two control points is large. We allow slight deviations by locally searching for a better box. We explicitly do not require this step to give us perfect boxes, but rather ``good enough" region proposals. We defer the final accuracy to an automatic refinement step, and our interactive segmentation annotation module. Note that in practice $N\ll M$, resulting in a huge saving in annotation time. We illustrate our interface in Fig.~\ref{fig:interface}.

We now describe how we obtain the curve for the object's trajectory. 
Our approach is optimization based: given the last annotated frame, we track the object using any of the existing trackers. In our work, we employ SiamMask~\cite{Tao_2016} for its speed and accuracy. We then take the tracked box and fit a polygonal curve with $N$ control points. 

Let $\cp_i=[t_i, x_i, y_i, w_i, h_i]^T$ denote the $i^{th}$ control point, where $t$ represents continuous time, and $[x_i, y_i,w_i,h_i]$ denotes the center of the box and its width and height, respectively. Let $P = \{\cp_1, \cp_1, \cdots, \cp_{N}\}$ to be the sequence of all control points.   
Note that our curve is continuous and parametric, \ie, $P(t)$, with $t_i\leq t\leq t_{i+1}$, we define $P(t) = [(t - t_i ) \cp_i + (t_{i+1} - t) \cp_{i+1}] / (t_{i+1} - t_i)$. 

\textbf{Curve Fitting:} To fit the curve to the observed boxes (obtained by the tracker), we initialize the control points by uniformly selecting key frames and placing points in the center of each frame, 
and run optimization. We uniformly sample $K$ points along both the parametric curve $P$ and the observed object track $J$, and optimize the following cost: \\
\begin{minipage}{\linewidth}
\begin{equation}
	L_{\mathrm{match}}(\{\cp_i\}) =\sum_{k=1}^{K} \lVert  [t_k^P, x_k^P, y_k^P, w_k^P, h_k^P]^T - [t_k^J, x_k^J, y_k^J, w_k^J, h_k^J]^T \rVert_1
\end{equation}
\end{minipage}
Note that each $[t_k^P, x_k^P, y_k^P, w_k^P, h_k^P]^T$ is a linear function of its neighboring control points, allowing us to compute the gradients with respect to  $\{{\cp_i}\}$. We use gradient descent to optimize our objective. 
 In our experiments, we choose $N$ = 10 and we set $K$ to be 300 to ensure number of sampled points is greater than number of frames. We run optimization for 100 steps.

\begin{figure*}[t!]
\centering
\includegraphics[width=\linewidth,trim=0 0 0 0,clip]{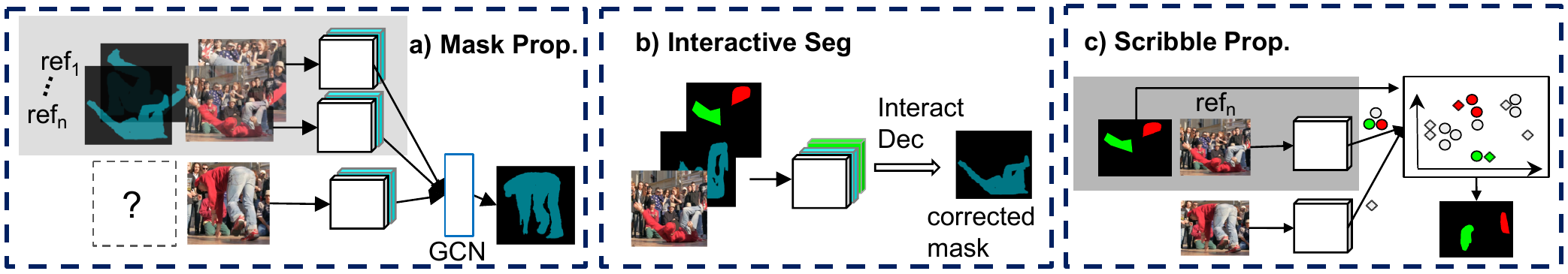}
\caption{\footnotesize  
Modules in our model:
{\bf a)} Given user annotated reference frames, our mask propagation module predicts object masks in subsequent frames. {\bf b)} Interactive segmentation module takes user scribbles as input (red and green scribbles indicate false positive and false negative regions, respectively), and returns the corrected mask for the chosen frame. {\bf c)} We propagate scribbles to correct masks in nearby frames.} 
\label{fig:interactive_seg}
\end{figure*}


\textbf{Interactive Annotation:} We pop the frames closest to the estimated control points, and allow the user to make adjustments if necessary. In particular, each control point can be adjusted both in space and time to accurately place its corresponding box around the object. The user is also asked to fast-forward the video between the control points, making sure that the object is loosely tracked in each interval. In case the interpolation loses the object, the annotator is able to add an additional control point. We then take the last annotated frame and re-run the tracker in the subsequent video. 
In our case, the last annotated box is used to compute correlation using SiamMask in the following frames. We re-fit the remaining control points to the subsequent track. The interactive process is repeated until the annotator deems the object to be well tracked.


\textbf{Box Refinement:} 
After interactive annotation, we get a curve which models piece-wise linear motions. To refine the linearly interpolated box in the frames in between every two control points, we crop the images based on the size of the box following SiamRPN~\cite{B_li18}  and re-run the SiamMask tracker by constraining it to these crops. This in practice produces significantly more accurate boxes that we exploit further in our interactive segmentation module. 

\subsection{Interactive Segmentation Annotation}
\label{sec:vos}
Given the annotated tracked box from Section~\ref{sec:VOT}, we now aim to interactively segment the object across the video with minimal human labour. We make use of the SiamMask's mask prediction in the first frame and predict masks for the rest of the video by our \emph{mask propagation module}. If errors occur, the user can edit the inaccurate masks by drawing positive and negative scribbles. Here, positive scribbles indicate missing regions, while negative scribbles indicate false positive regions. To best utilize human's feedback, our method propagates both the corrected mask as well as the input scribbles to the subsequent frames. We refer to our interactive segmentation module as \emph{Scribble-VOS}.

\subsubsection{Network Design}

Our interactive segmentation approach consists of three different modules: a \emph{mask propagation module} to propogate segmentation masks, an \emph{interactive segmentation module} for single image editing via scribbles, and a \emph{scribble propagation module} that propagates user's feedback across video. Overview is in Fig~\ref{fig:interactive_seg}. We explain each module in detail next.  

\textbf{Image Encoder:}
We crop the image inside the box and encode it with ResNet50~\cite{he2016deep} pre-trained on ImageNet. We extract image features from the Conv4 layer and denote it as $F_t\in \mathbb{R}^{W\times H\times D}$, where $t$ is the time step in the video, and $W$ and $H$ are the width and height of the feature map. Note that unlike most previous works that concatenate images with additional information including masks/scribbles as input to the encoder, our encoder only takes images as input, which allows us to share the encoder among different modules and use an encoder pre-trained on any (large) image dataset. 

\textbf{Mask Decoder:} There are two decoders in our framework.
\textit{Interactive decoder} is used to produce a refined mask of single frame based on the human interaction. We also have a \textit{propagation decoder} which is shared by both the mask and scribble propagation modules. 
We describe the structure of the decoder next, and specify the input feature for the two decoders later in the section. 

Our decoder consists of three refinement heads~\cite{wug2018fast} that upsample  image features (1/16 resolution wrt input) by 8x to produce the object mask. Sizes of the output channels of the refinement heads are 224, 224, 128. We also remove the skip connection for the last refinement head as it yields  better results. 


\textbf{Interactive Segmentation Module:}
To correct errors produced by the model, our interactive module takes user's scribbles for the chosen frame and outputs a refined mask. 
Specifically, we first convert user's scribbles into two binary maps, one for positive and the other for negative scribbles. 
The input to the \emph{interactive decoder} concatenates 2-channel scribble input, image features, and the mask of the frame that we want to correct, as shown in Fig.~\ref{fig:interactive_seg}b.

To force the model to behave consistently with the user's scribbles, we utilize a \emph{scribble consistency loss}. And, to prevent the predicted mask from being modified in regions where the initial mask was accurate, we restrict the network to only affect a local neighbourhood surrounding the scribbles. In particular, we mask out new predictions that are more than 10 pixels away from the scribble areas. 

Note that in~\cite{oh2019fast}, both scribbles and masks are encoded along with the raw image, which requires an additional encoding for each interaction. Here, we only run the encoding once and re-use the image feature for each interaction.  

\textbf{Mask Propagation Module:}
Given the user-annotated object masks, our mask propagation module aims at re-predicting the masks for the subsequent frames. We rely on a Graph Convolutional Network (GCN)~\cite{DBLP:conf/iclr/KipfW17} to perform this propagation, and additionally employ a decoder which produces the segmentation mask based on the GCN features. 
Let $R$ denotes the set of user-annotated frames, which serve as reference frames for propagation, and let $c$ denotes the current frame that we wish to predict the mask for. We build a graph $G=(V, E)$, where $V$ denotes nodes and $E$ denotes edges to encode the structure between frames in $R$ and the current frame $c$. In particular, we make every location in the $W \times H$ feature map of every frame a node in the graph, \ie, there are $(|R|+1)\times W \times H$ nodes in total. Each node in frame $c$ is connected to all the nodes in frames $\in R$ via an edge. An example of the graph is illustrated in Fig.~\ref{fig:interactive_seg}a.

As input to our GCN,  we concatenate the image feature with corresponding masks for each frame in $R$. We perform a similar concatenation operation for $c$ but use a uniform ``mask'' instead. 
The input feature $f_u$ for vertex $u$ in GCN is computed as $f_u = \text{concat}\{F(x, y), M(x, y)\}\in \mathbb{R}^{(D+1)}$, where $F$ and $M$ are the image feature and the corresponding mask at time step $t_u$, respectively, and $D$ is dimension. Here, $(x, y)$ is the coordinate of the node $u$ in the feature map. 

We assign a weight $w_e(u, v)=\frac{\exp(f_u\cdot f_v)}{\sum_{v^{\prime}\in \mathcal{N}(u)}\exp(f_u\cdot f_{v^{\prime}})}$ to each edge  $(u, v)$ in the graph. 
The following propagation step is performed for node $u$:\\
\begin{minipage}{\linewidth}
\begin{eqnarray}
f_u^{\prime} = W_0 f_u+\sum_{v\in \mathcal{N}(u)}w_e(u, v) W_1 f_v,
\end{eqnarray}
\end{minipage}
where $\mathcal{N}(u)$ denotes neighbours of $u$ in the graph, and $W_0, W_1$ are the weight matrices of the GCN layers. 
We take the feature $f_u^{\prime}\in \mathbb{R}^K$ for all nodes $u$ in current frame $c$ as the input feature $F_c^g \in \mathbb{R}^{W\times H\times D^\prime}$ to the \emph{propagation decoder} which predicts the refined mask. We use $D^\prime$ to denote the output dimension of the GCN feature, and it is also the dimension of the input feature for the decoder. The superscript $g$ stands for GCN. The feature, $F_c^g$, is used by the Scribble Propagation module described next. 

\textbf{Scribble Propagation Module:}
Most existing works employ user's corrections by only propagating the annotated masks to other frames. 
We found that propagating the scribbles explicitly produces notably better results. 
By feeding scribble information explicitly to the network, we are providing useful information about the regions it should pay attention to. Qualitatively, without providing this information (see Fig.~\ref{fig:ipn_ours_comparison}), we notice that the propagation module typically ignores the corrections and repeats its mistakes in the subsequent frames. 

As shown in Fig.~\ref{fig:interactive_seg}c, inspired by~\cite{chen2018blazingly}, we formulate scribble propagation problem as a pixel-wise retrieval problem. For each pixel in current frame $c$, we find the pixel in the reference frame with the most similar embedding and assign the label of that pixel to the pixel in $c$. In the scribble propagation stage, we choose the reference frame as the annotated frame that is closest to the current time step. 
Formally, we first project the encoded pixel features into an embedding space and adopt the label of the nearest neighbour in this space. 
Pixels in the reference frame can be classified into three classes: background (regions with no scribbles), negative, and positive scribbles, respectively. 
We first project image features $F \in \mathbb{R}^{W\times H\times D}$ of the reference and current frames into a $d<D$ dimensional space using a shared embedding head. We use $d=128$ and a $3\times 3$ convolutional layer for the projection. 
We denote the transferred label for frame $c$ as $S_c \in\mathbb{R}^{W\times H\times 2}$. 
To get the final mask, we concatenate $S_c $ with $F^g_c$, \ie, the feature obtained from the GCN, reduce the dimension of the concatenation to $D^{\prime}$, and then feed it to the \emph{propagation decoder}. 
Here, we employ a $3\times 3$ convolution layer to perform the dimension reduction.
Thus, the input feature to the \emph{propagation decoder} is
\begin{equation}
F_c^{gs}=M(\mathrm{concat}\{F_c^g, S_c\})\in\mathbb{R}^{W\times H\times D^{\prime}},
\end{equation}
where $M$ is the conv layer for dimensionality reduction.
Superscript $gs$ denotes merging GCN's output and the predicted scribbles. 
The \emph{propagation decoder} is shared for mask and scribble propagation. The decoder takes the encoded feature (either the feature $F_c^g$ from the mask propagation module or the feature $F_c^{gs}$ from the scribble propagation stage), and outputs an object mask. 
This scribble network propagation is designed for propagating local errors within a short range and thus is different in purpose from the mask propagation module.



\subsubsection{Network Training:}
\label{sec:NetworkTraining}

All networks are trained end-to-end using binary cross entropy loss for mask prediction. To better supervise our model, we further use Scribble Consistency Loss for training the \emph{interactive segmentation module}, and Batch Hard Triplet Loss~\cite{chen2018blazingly} for training the \emph{scribble propagation module}. We first describe how we synthesize scribbles and then specify the scribble consistency loss in the following paragraphs. We defer the Batch Hard Triplet Loss and implementation details to the appendix.

\textbf{Scribble Correspondence:} 
A big challenge of training the scribble propagation module is the lack of ground-truth pixel-wise correspondences between the scribbles in frame $r$ and scribbles in frame $c$.
To solve this problem, we create the ground-truth correspondences by synthesizing new frames: we augment the reference frame $r$ to obtain the current frame $c$. The same augmentation is applied on the scribble map which gives us ground-truth correspondences between scribbles. Specifically, we use thin-plate spline transformation with 4 control points for the augmentation. 

\textbf{Scribble Consistency Loss:}  Typically, a user, who draws the scribble, would expect the network to behave consistently with the annotation, \ie, expects to see positive predictions around positive scribbles and vice versa for the negative. To encourage this behavior, we employ a scribble consistency loss:
\begin{equation} 
L_{\mathrm{sc}} = \sum_{i,j}-S_{\mathrm{p}}(i,j)\log M_{\mathrm{pred}}(i,j)
-S_{\mathrm{n}}(i,j)\log (1-M_{\mathrm{pred}}(i,j)),
\end{equation} 
where $S_{\mathrm{p}}$ and $S_{\mathrm{n}}$ are the positive and negative scribble maps, $(i, j)$ is the coordinate on the mask. Here, $M_{\mathrm{pred}}$ denotes the predicted object mask. Without this loss, the model in many cases ignores the user's input. 

\begin{table*}[t!]
\begin{minipage}{0.58\linewidth}
\begin{center}
{
\addtolength{\tabcolsep}{1.9pt}
\begin{footnotesize}
\resizebox{\columnwidth}{!}{
\begin{tabular}{|l|c|c|c|c|}
\hline
Correction Rounds & 0 & 4 & 6 & 9 \\ 
\hline
\begin{tabular}[c]{@{}l@{}} IPN-Box \& Line-Scrib. \end{tabular} & 76.93 & 78.89 & 79.73 & 80.88 \\ 
\hline
\begin{tabular}[c]{@{}l@{}} GCN Mask Prop. \& Line-Scrib.
\end{tabular} 
&79.33 & 86.70 & 87.48 & 88.48 \\ 
\hline
\begin{tabular}[c]{@{}l@{}}  GCN Mask Prop. \& Area-Scrib.
\end{tabular}& 79.33 & 87.62 & 89.07 & 89.75 \\ 

+ Scribble Prop. & - & 88.92 & 90.90  & 91.16 \\ 
+ GT First Frame & 85.14 & 89.61 & 90.28  & 90.91 \\ 
\hline
\end{tabular}
}
\end{footnotesize}
\caption{\footnotesize {\bf Ablation study} (DAVIS'17). Round 0 means mask prop. with predicted first frame mask.}
\label{tbl:Ablation}
}

\end{center}
\end{minipage}
\hfill
\begin{minipage}{0.39\linewidth}
\begin{center}
{
\addtolength{\tabcolsep}{0pt}
\resizebox{\columnwidth}{!}{
\begin{tabular}{|c|c|c|c|}
\hline
Model & IPN & Ours-256 &  Ours-512\\ 
\hline 
Interaction &  18  &  10.5 & 12 \\
\hline
Curve Fitting & - &  12 & 12 \\
\hline
Mask Prop.  & 15 &  29&  54\\ 
 w/Cache  & - & 8 & 26\\
\hline 
Scribble Prop.  & -  & 15 & 28\\
\hline
\end{tabular}
}
\caption{\footnotesize {\bf Running Time}. Numbers are reported in ms per frame.}
\label{tbl:running_time}
}
\end{center}
\end{minipage}
\end{table*}
\begin{figure*}[t!]
\addtolength{\tabcolsep}{-1pt}
\begin{tabular}{ccccc}
\includegraphics[width=0.2\linewidth,trim=0 15 0 35,clip]{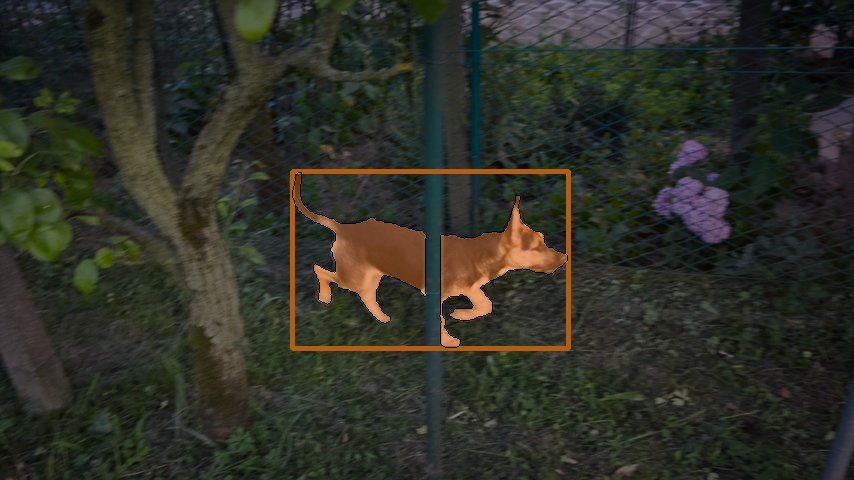} &
\includegraphics[width=0.2\linewidth,trim=0 15 0 35,clip]{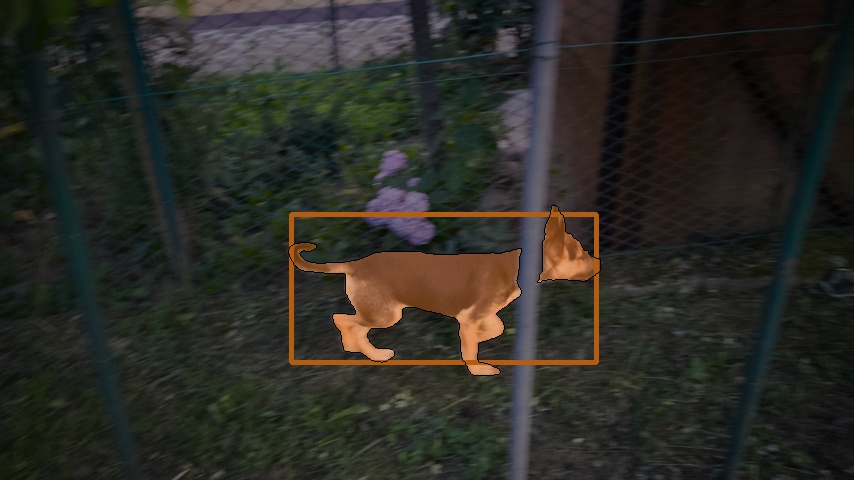} &
\includegraphics[width=0.2\linewidth,trim=0 15 0 35,clip]{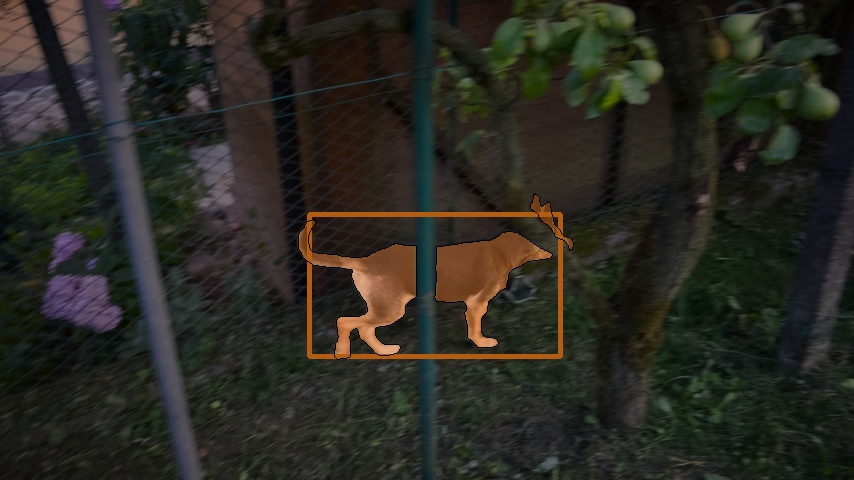} &
\includegraphics[ width=0.2\linewidth,trim=0 15 0 35,clip]{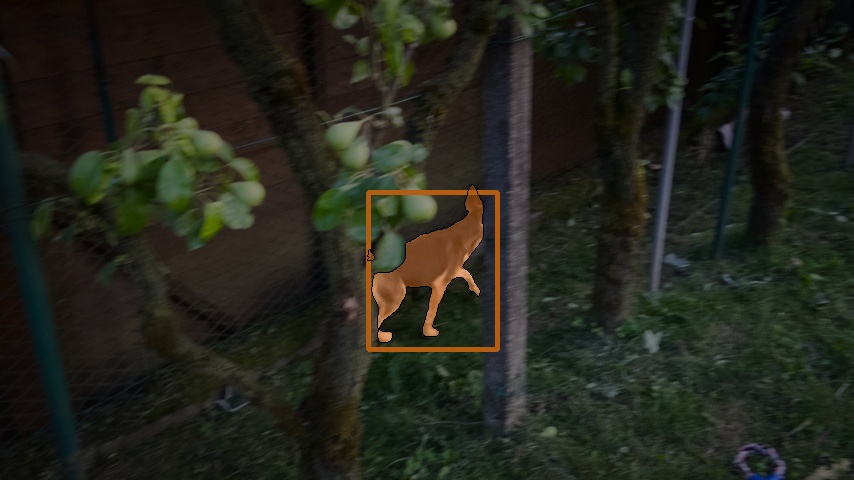} &
\includegraphics[width=0.2\linewidth,trim=0 15 0 35,clip]{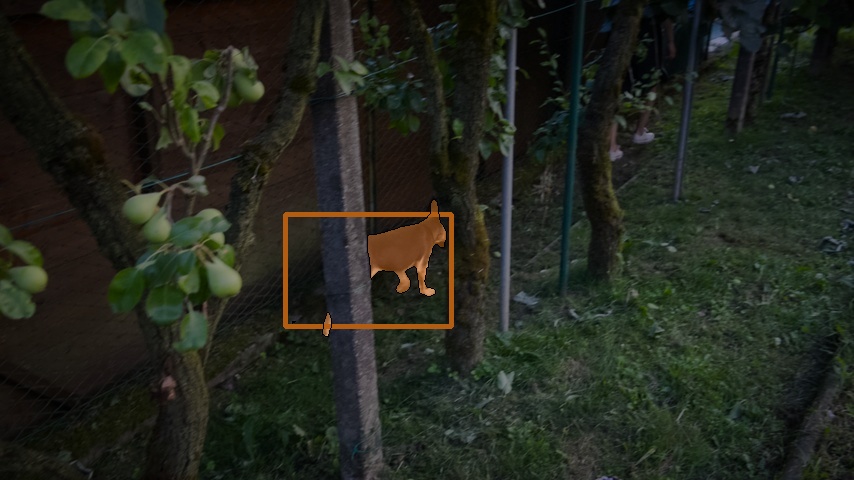} \\[-0.6mm]
\includegraphics[ width=0.2\linewidth,trim=0 0 0 25,clip]{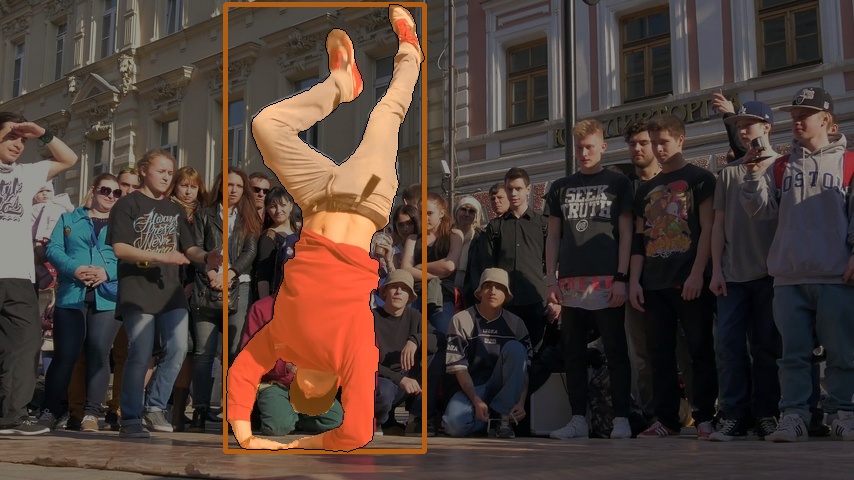} &
\includegraphics[width=0.2\linewidth,trim=0 0 0 25,clip]{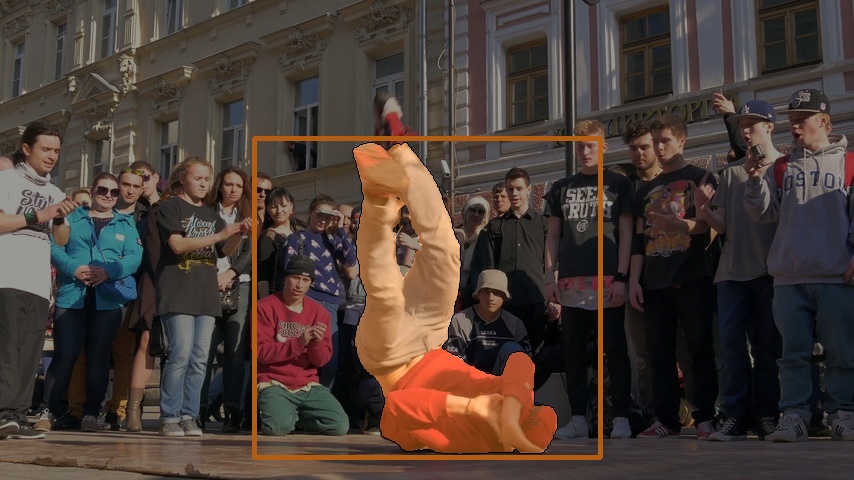} &
\includegraphics[width=0.2\linewidth,trim=0 0 0 25,clip]{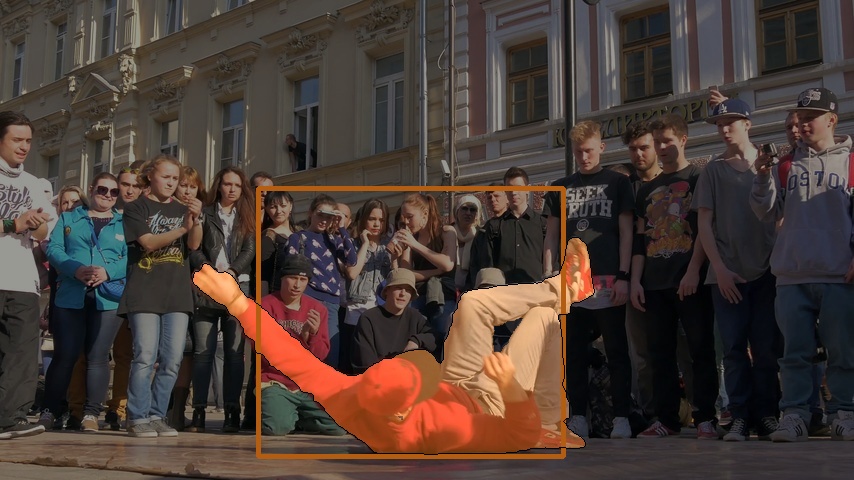} &
\includegraphics[width=0.2\linewidth,trim=0 0 0 25,clip]{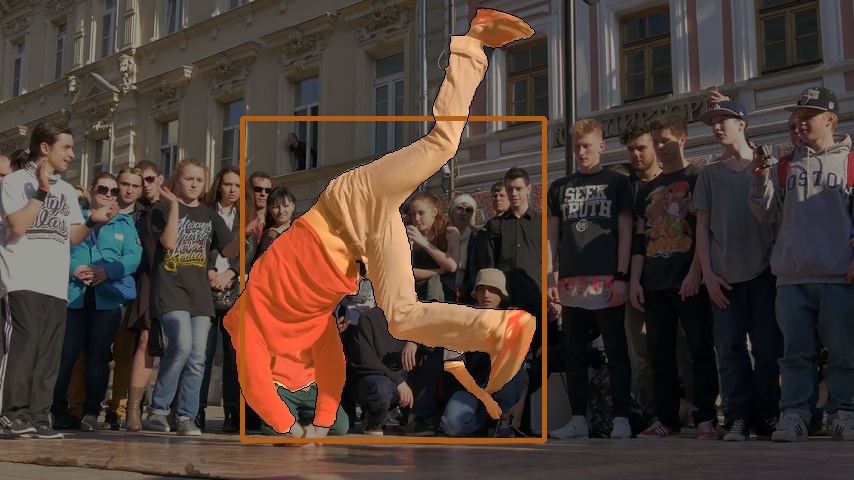} &
\includegraphics[width=0.2\linewidth,trim=0 0 0 25,clip]{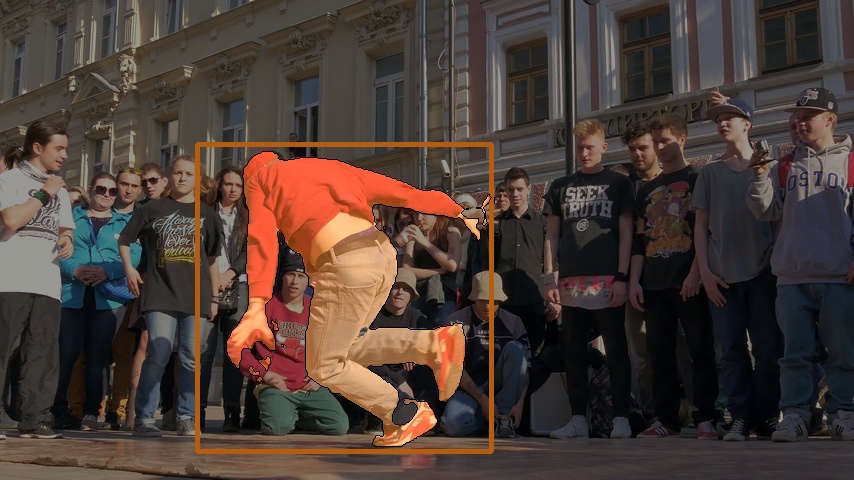}  \\[-0.6mm]
\includegraphics[width=0.2\linewidth,trim=0 0 0 25,clip]{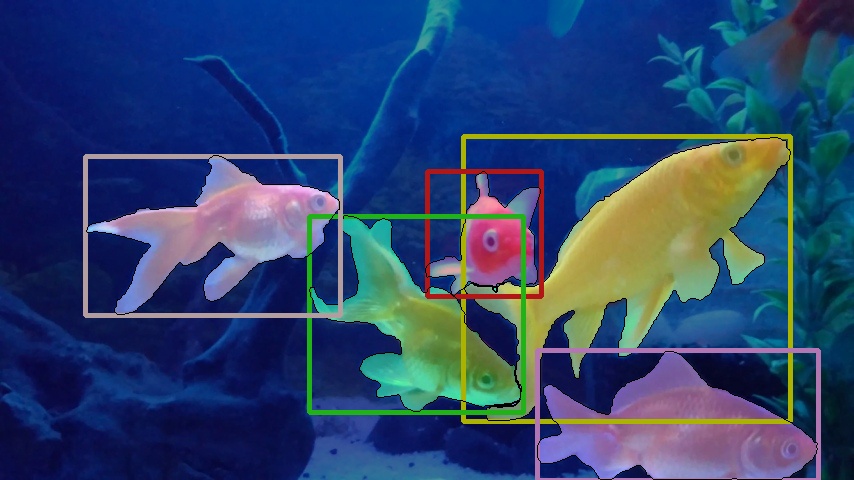} &
\includegraphics[width=0.2\linewidth,trim=0 0 0 25,clip]{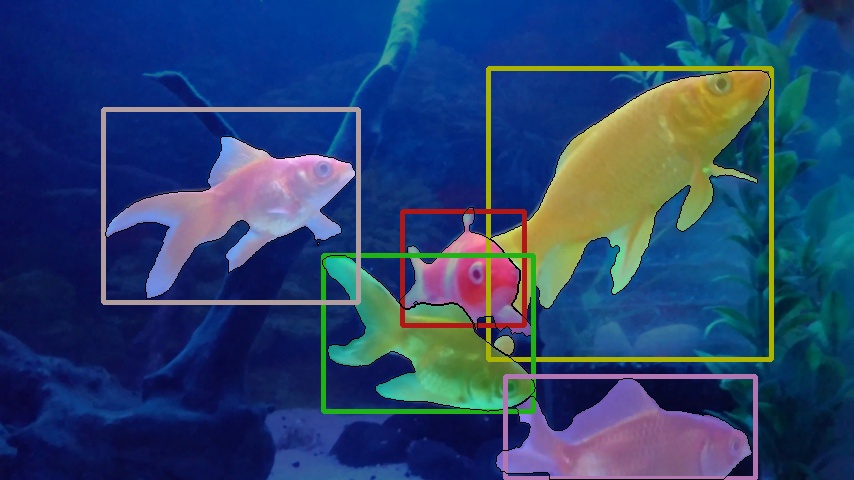} &
\includegraphics[width=0.2\linewidth,trim=0 0 0 25,clip]{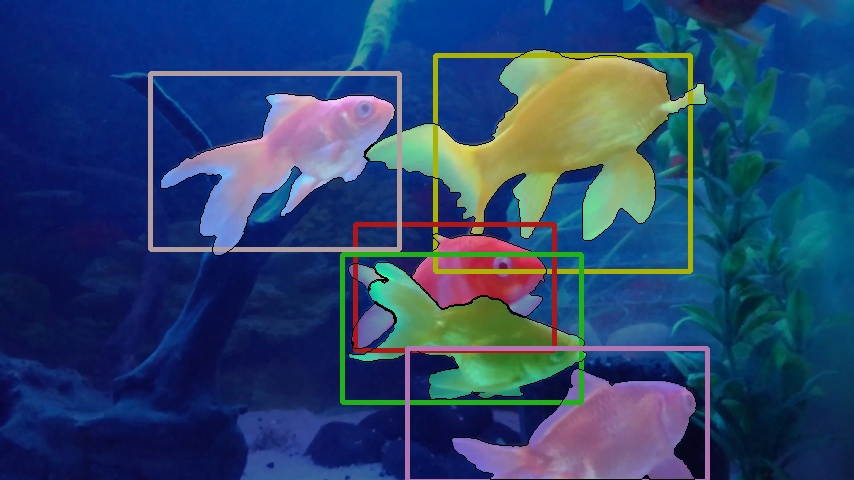} &
\includegraphics[width=0.2\linewidth,trim=0 0 0 25,clip]{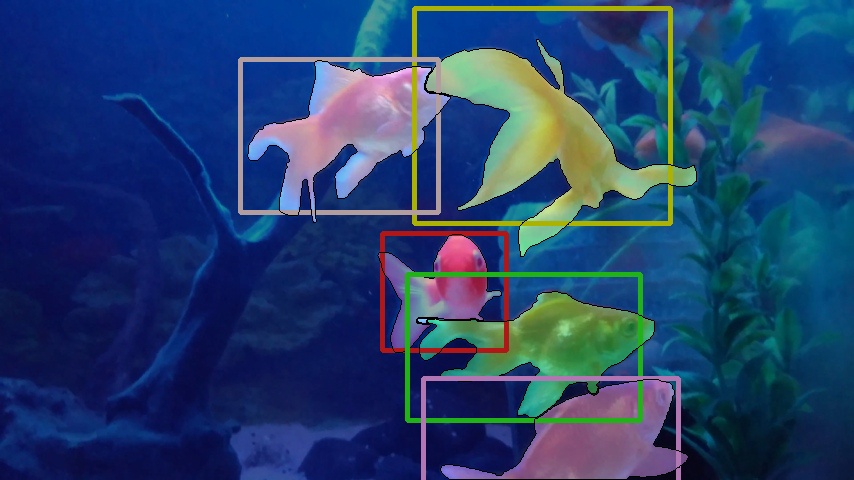} &
\includegraphics[width=0.2\linewidth,trim=0 0 0 25,clip]{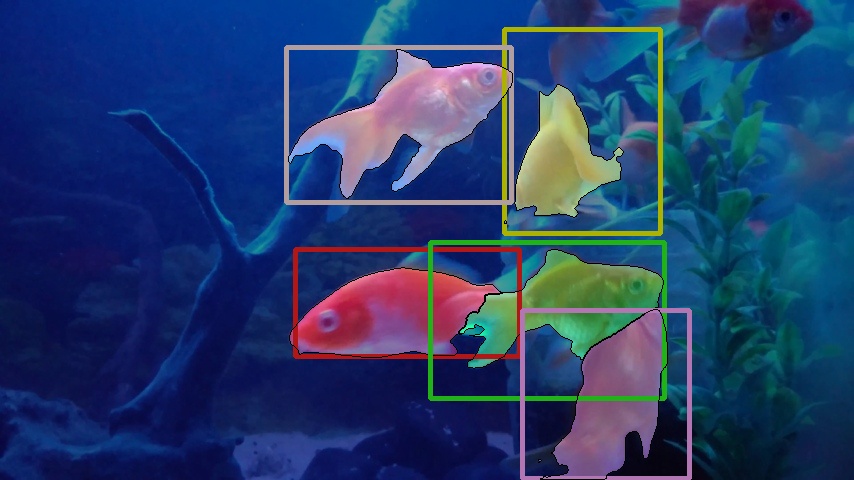} \\
\end{tabular}
\caption{\footnotesize Qualitative results on DAVIS2017 val set using our full annotation framework. 
 5 rounds of scribble correction + 8.85 box corrections were used.}
\label{fig:results_DAVIS}
\end{figure*}
\section{Experimental Results}
\label{sec:results}
We perform extensive evaluation of ScribbleBox for video object annotation. To test generalization capabilities, we evaluate our approach both in the same domain, as well as in the out-of-domain datasets. For the In-Domain experiment, following IPN~\cite{oh2019fast}, we show results on DAVIS2017~\cite{KhoRohrSch_ACCV2018} validation set, which has 65.3 frames per object on average. For the Out-of-Domain experiment, we showcase our results on MOTS-KITTI~\cite{voigtlaender2019mots} which is a recent multi-object tracking and segmentation subset of KITTI~\cite{KITTI}. We evaluate on a subset of 514 objects from the training and validation set where video length is between 20 and 150 frames. 
The average video length is 48.8 frames per object. We also perform a user study with real annotators labeling videos with our annotation tool. 

\textbf{Baseline:} We use IPN~\cite{oh2019fast} as our baseline, which ranks first in DAVIS'18 Interactive Challenge. Note that the official benchmark is not applicable in our case since our method is a two-stage method and the DAVIS-Agent does not support bounding box correction. Thus, we evaluate against the baseline by reporting box correction and scribble drawing effort v.s J\&F via simulation (Table~\ref{tbl:Ablation}, Fig.~\ref{fig:full_pipline}) and annotation time v.s J\&F through a user study (Fig.~\ref{fig:userstudy}).

\textbf{Scribbles:}
DAVIS Interactive Challenge~\cite{Caelles_arXiv_2018} defines scribbles as thin skeletons inside erroneous areas, which we refer to as Thin-Line-Scribble. We instead propose to use more informative scribbles, where we ask the user to roughly trace the erroneous area with a cursor. We name our scribbles as Area-Scribble. While this may require slightly more effort, we show significant performance gains in our ablation study for both types of scribbles. 

To simulate the Area-Scribble, given a previously predicted object mask and the ground-truth mask, we sample positive and negative scribbles from the false negative and false positive areas, respectively. To simulate realistic ``trace'' scribbles that a human would provide, we perform a binary erosion to the erroneous area and take the resulting regions as our simulated scribbles.



\textbf{Details:} 
We use the official SiamMask~\cite{Siammask} tracker provided by the authors which was trained on COCO~\cite{Lin_2014},
ImageNet\-VID~\cite{ILSVRC15} and YouTube\-VOS~\cite{Xu_2018}. We perform inference for each object in the video. 
In Scribble-VOS, we propagate the corrected mask using mask propagation network to the end of the video. In addition, 
we propagate scribbles in two directions (forward and backward) for at most $n$ frames (we use $n=5$ in experiments).

\begin{figure*}[b!]
		\centering
\begin{minipage}{0.62\linewidth}
	\begin{minipage}{0.49\linewidth}
		\includegraphics[width=0.99\linewidth,trim=10 26 0 0,clip]{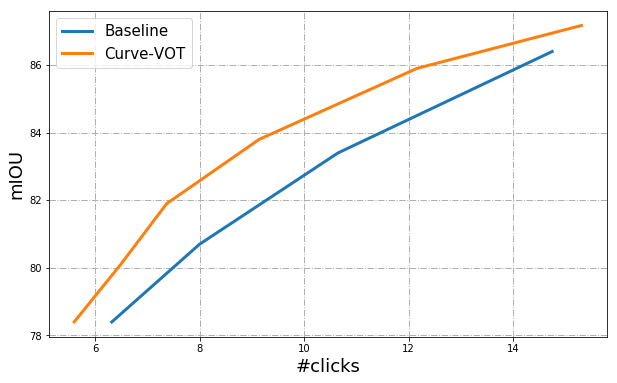}\\[-1mm]
		\includegraphics[width=0.99\linewidth,,trim=10 0 0 0,clip]{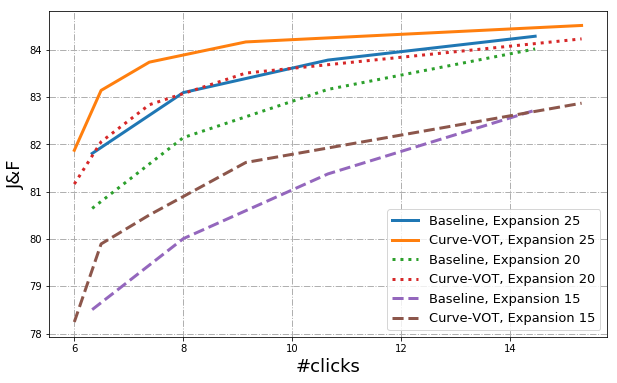}
	\end{minipage}
	\begin{minipage}{0.49\linewidth}

		\includegraphics[width=0.99\linewidth,trim=10 26 0 0,clip]{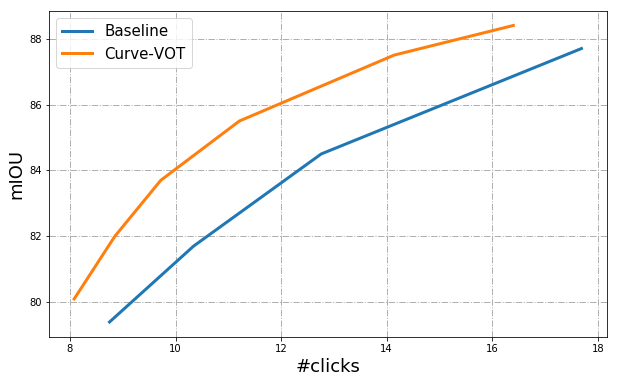}\\[-1mm]
		\includegraphics[width=0.99\linewidth,trim=10 0 0 0,clip]{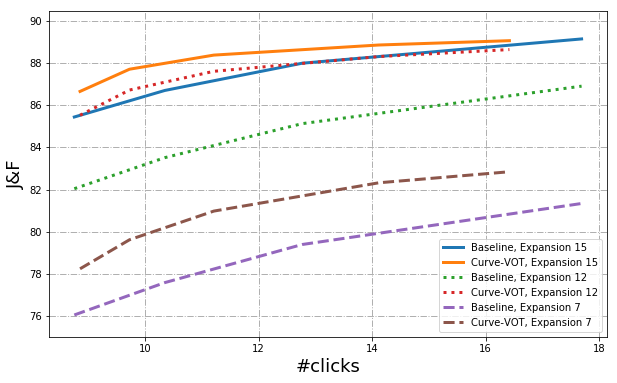}
	\end{minipage}
	\begin{center}
		\begin{tabular}{cc}
		\scriptsize{(DAVIS2017)} $\qquad\qquad\qquad \ $ & \scriptsize { (KITTI)}\\
	\end{tabular}
	\end{center}
	\end{minipage}
\begin{minipage}{0.36\linewidth}
	\caption{\footnotesize {\bf Interactive-VOT}:  (top) Box IoU vs number of simulated clicks (we report IoU averaged across frames), (bottom) Segmentation J\&F. Note that in this experiment segmentation is automatic using annotated boxes, i.e., there are no clicks to improve masks, only to improve boxes.}
	\label{fig:bbox_davis}
\end{minipage}
\end{figure*}

\subsection{In-Domain Annotation}

\textbf{Tracking Annotation:}  
We compare Curve-VOT with a brute-force baseline which also uses SiamMask~\cite{Siammask}. Instead of manipulating trajectories via curves, the user watches the video and corrects a box when it significantly deviates from the object. We use this correction as a new reference and run the tracker again starting from this frame. We simulate the user by correcting frames with IOU lower than a threshold, where we evaluate different thresholds from 0.4 to 0.8, representing different effort levels by the users. We treat each user's correction as two clicks, i.e., representing two corners of the box. 

We first compare our approach with the baseline in terms of box IOU. As shown in the top-left of Fig.~\ref{fig:bbox_davis}, our Curve-VOT requires fewer clicks to achieve the same IOU. In the bottom-left of Fig.~\ref{fig:bbox_davis}, we provide a comparison in terms of segmentation J\&F. We take the corrected bounding boxes and expand them to crop image. We report results with different expansion sizes. 
Since tracking the box only acts as a region proposal for VOS, we evaluate tolerance for tracking noise. As shown in the bottom-left of Fig.~\ref{fig:bbox_davis}, using a 25 pixels expansion, J\&F converges at around 9 clicks. This plot also emphasizes the importance of Scribble-VOS since additional box clicks only increase the accuracy marginally, which is a wasted effort. We instead turn to Scribble Annotation to further improve results. 

\begin{figure*}[t!]
\begin{minipage}{0.64\linewidth}
\addtolength{\tabcolsep}{-3.2pt}
\begin{tabular}{cc}
\includegraphics[width=0.49\linewidth,trim=4 0 0 0,clip]{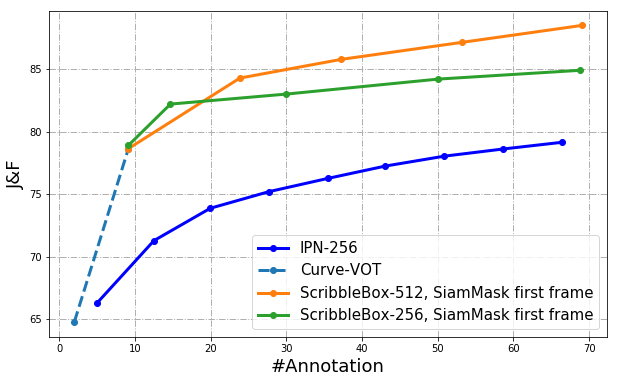}&
\includegraphics[width=0.49\linewidth,trim=4 0 0 0,clip]{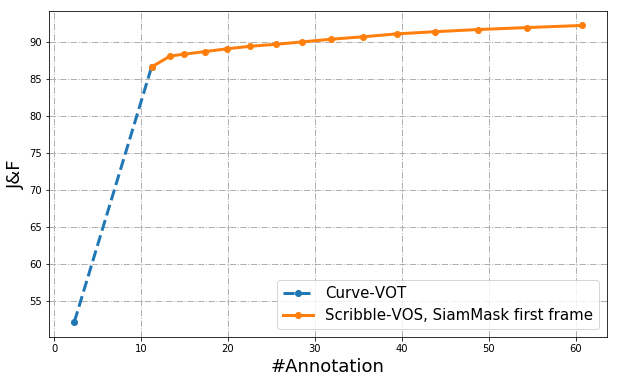} 
\end{tabular}
\caption{\footnotesize {\bf Simulated} full annotation workflow on (left) DAVIS2017, (right) KITTI. Dashed blue lines denote Curve-VOT, continuing lines Scribble-VOS annotation.}
\label{fig:full_pipline}
\end{minipage}
\hfill
\begin{minipage}{0.324\linewidth}

\includegraphics[width=1\linewidth,trim=4 0 0 0,clip]{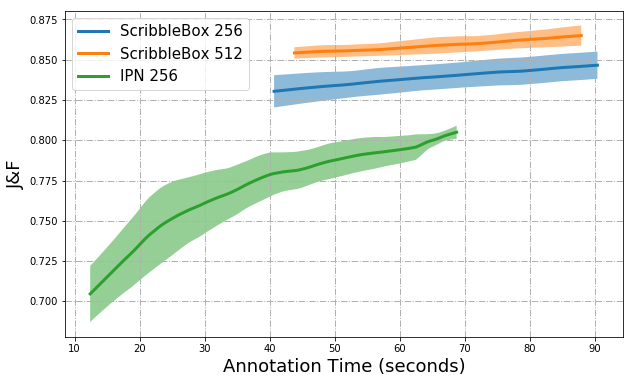} \\[-7mm]
\vspace{3.8mm}
\caption{\footnotesize Real annotator {\bf user study} on DAVIS'17. Filled area denotes variance.}
\label{fig:userstudy}
\end{minipage}
\end{figure*}

\textbf{Mask Annotation:}  We first conduct an ablation study on individual components in Scribble-VOS in Table~\ref{tbl:Ablation}.  Following IPN~\cite{oh2019fast}, we use the ground truth box in the first frame. To remove the effect of box annotation, we modify the IPN baseline to work on boxes.  We name this baseline as IPN-box, where we use the officially released code from~\cite{oh2019fast} and take box-cropped image as input. Same as our model, we first pre-train IPN-box on synthetic video clips, then fine-tune it on DAVIS2017. We add GCN-Mask Propagation module and multi-reference frames for interactions to the IPN-box baseline. It gives 2.4\%, 7.81\%, 7.75\%, 7.6\% improvements, respectively. To ablate our human interaction, we replace Thin-Line-Scribble with Area-Scribble. Our Area-Scribble yields 0.92\%, 1.59\% and 1.33\% improvements, respectively. Results of ScribbleBox without scribble propagation also indicate that our \emph{scribble propagation module} plays an important role in improving J\&F (1.3\%, 1.83\%, and 1.41\%, respectively). 
We further analyze impact on quality of the first frame's mask. As shown in the last line of Table~\ref{tbl:Ablation}, ground truth first frame mask gives 5.81\% improvement without interactions but the improvement becomes marginal after more rounds of interaction. These demonstrate the effectiveness of our interaction model. 

\textbf{Full Framework Comparison:} We conduct a full framework analysis in a simulated environment.  
We compare with previous work in terms of the number of human interactions. For a baseline, we run the officially released DAVIS Interactive agent locally and evaluate upon IPN's checkpoint which achieved the first place in DAVIS2018 Interactive Challenge. Each correction round returned by the DAVIS-Agent consists of multiple scribbles. We count each scribble as one annotation. Our results are conducted on the annotated tracked boxes after 9.14 box clicks (J\&F 78.64\%, \ie, second point on the dashed line in Fig.~\ref{fig:full_pipline}). Fig.~\ref{fig:full_pipline} reports a comparison of our framework with IPN~\cite{oh2019fast}. Note that our model with only Curve-VOT interaction  already performs better than baseline with mask annotation. We also ablate the impact of different image resolutions.

\begin{figure*}[t!]
\addtolength{\tabcolsep}{-0.8pt}
\begin{tabular}{cccccc}
\includegraphics[width=0.163\linewidth, trim=40 15 10 30,clip]{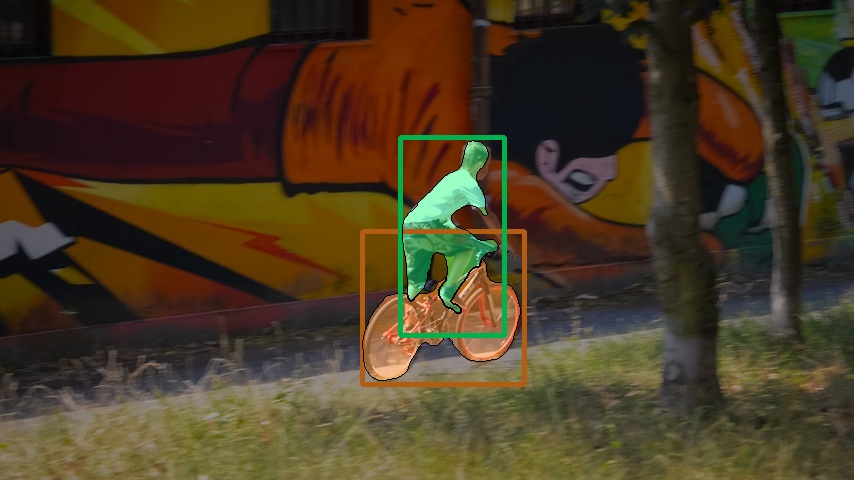} & 
\includegraphics[width=0.163\linewidth, trim=40 15 10 30,clip]{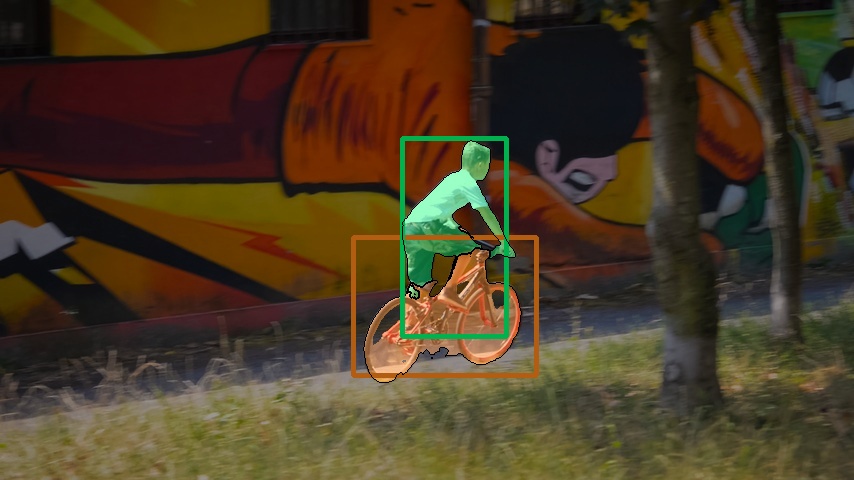} &
\includegraphics[width=0.163\linewidth, trim=40 15 10 30,clip]{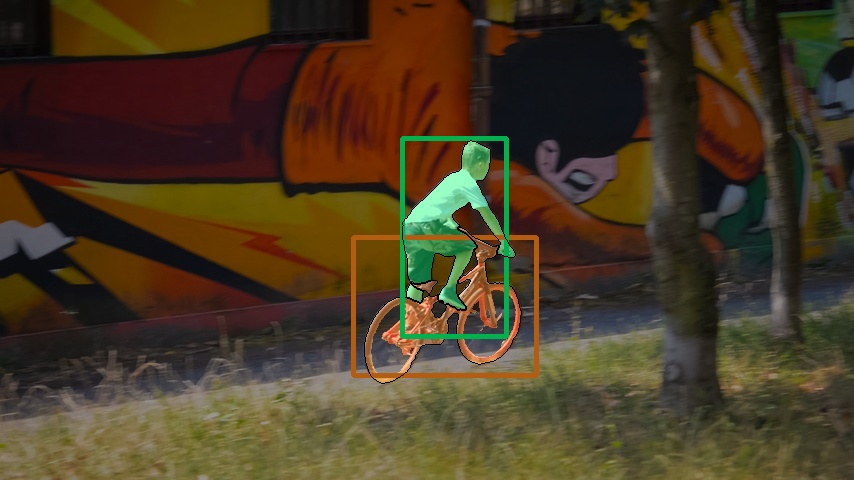} & 
\includegraphics[width=0.163\linewidth, trim=40 15 10 30,clip]{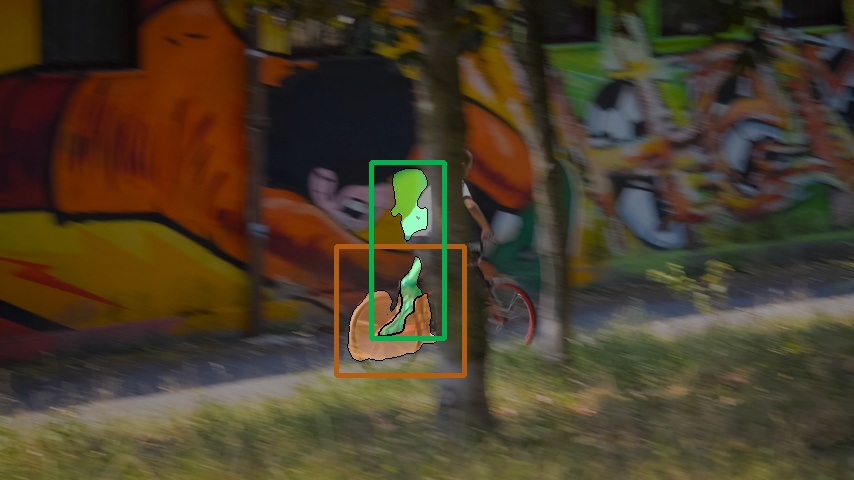} &
\includegraphics[width=0.163\linewidth, trim=40 15 10 30,clip]{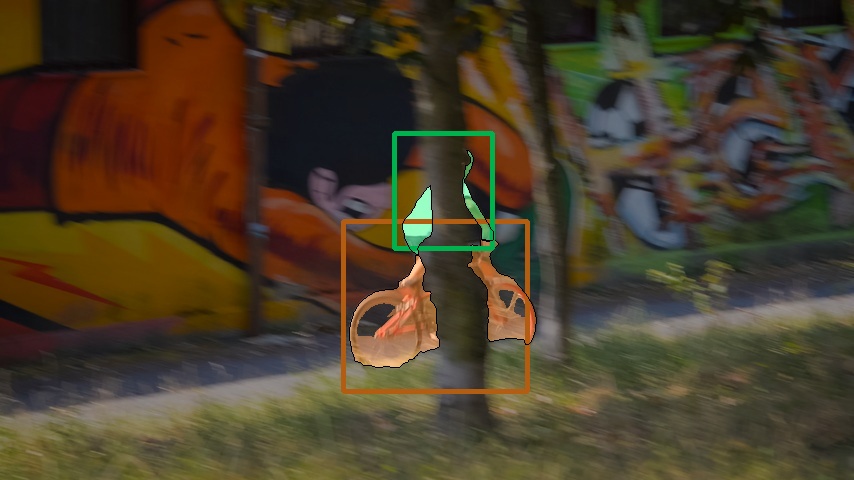} &
\includegraphics[width=0.163\linewidth, trim=40 15 10 30,clip]{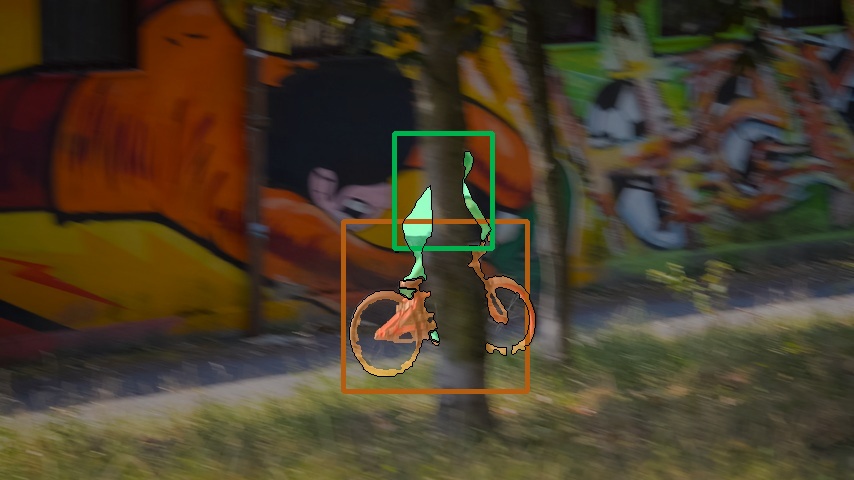}  \\[-0.6mm]
\includegraphics[width=0.163\linewidth, ,trim=40 0 10 0, clip]{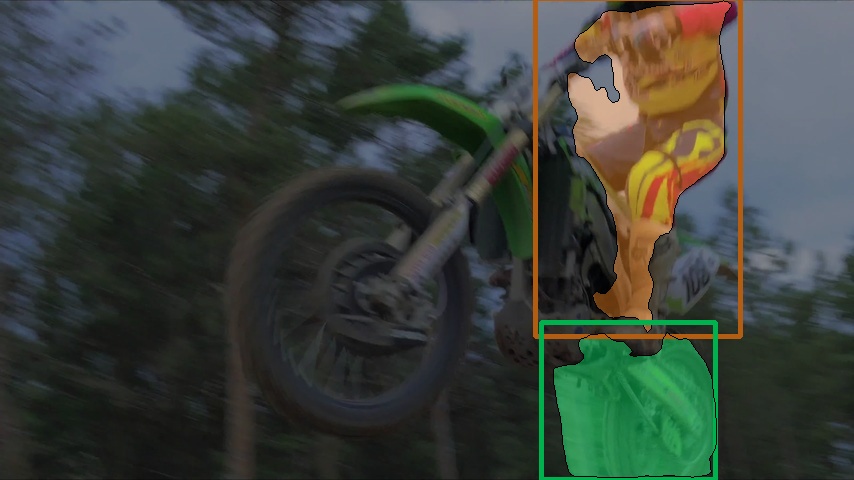} &
\includegraphics[width=0.163\linewidth,trim=40 0 10 0, clip]{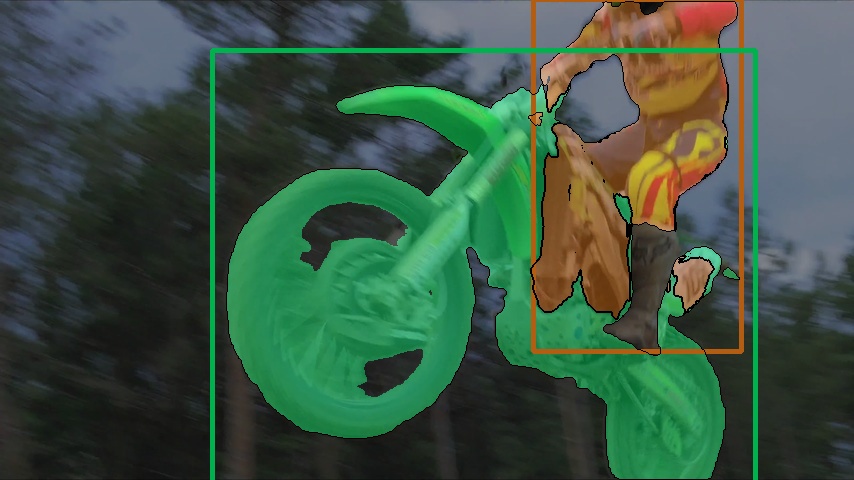} &
\includegraphics[width=0.163\linewidth, trim=40 0 10 0, clip]{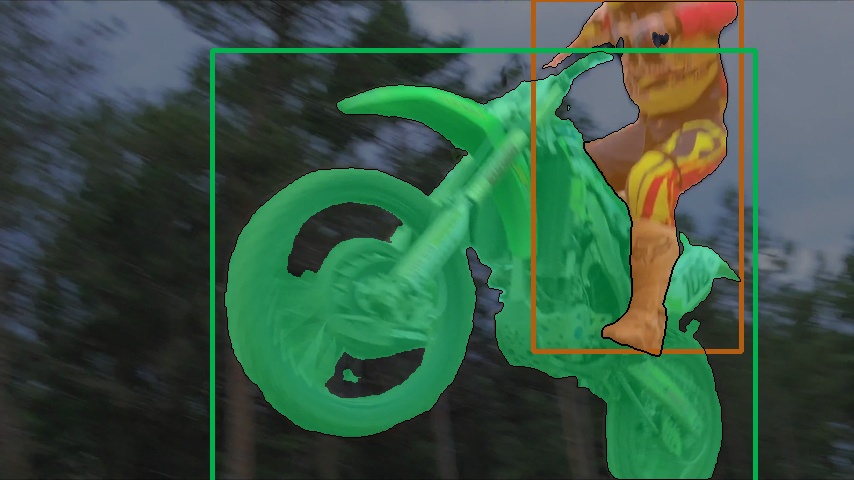} &   
\includegraphics[width=0.163\linewidth, trim=40 0 10 0, clip]{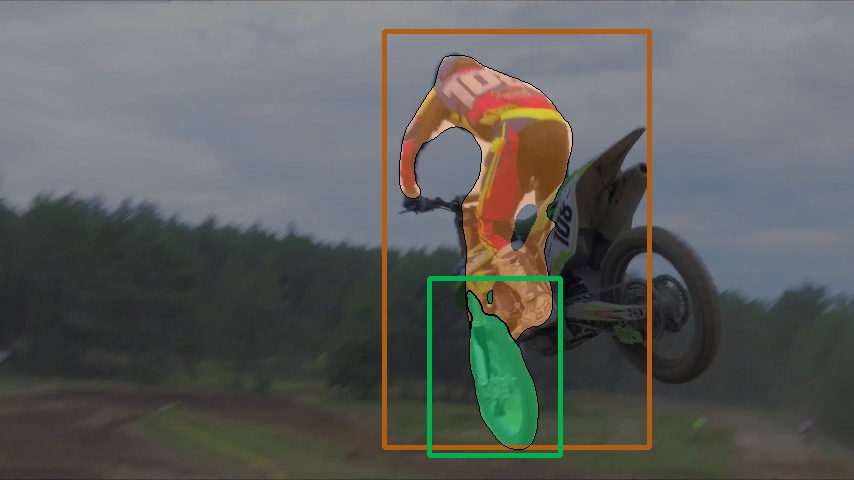} &
\includegraphics[width=0.163\linewidth, trim=40 0 10 0, clip]{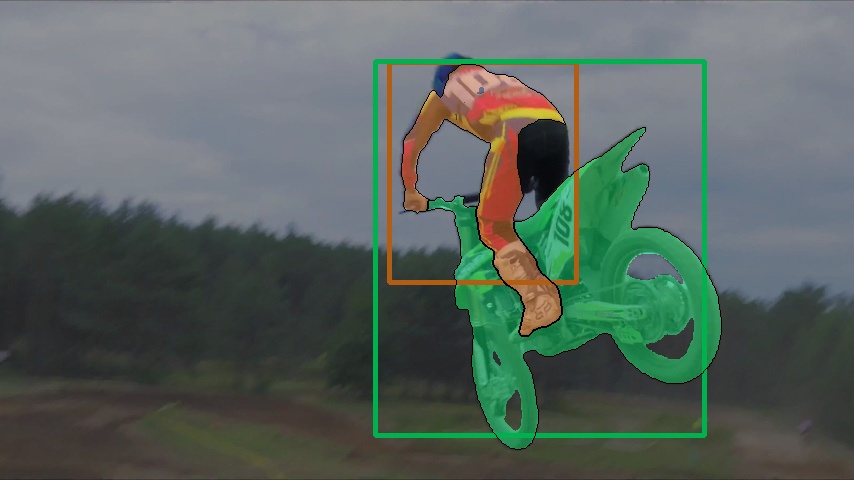} &
\includegraphics[width=0.163\linewidth, trim=40 0 10 0, clip]{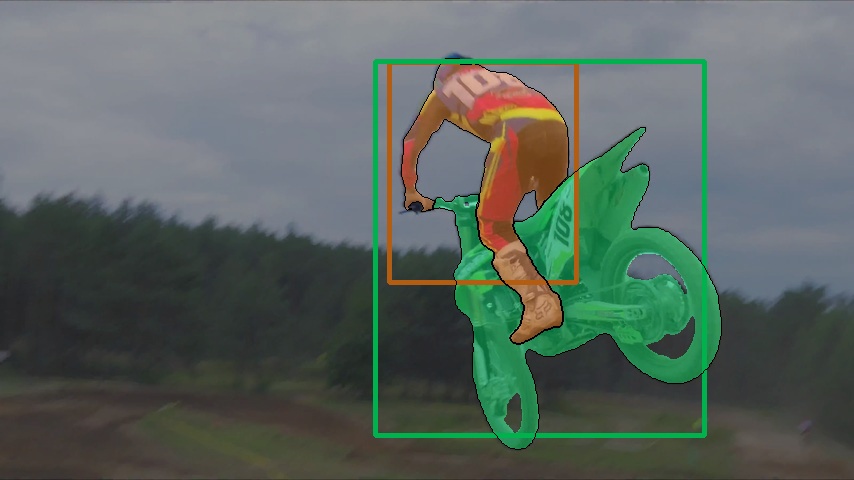}  \\[-1mm]
{\scriptsize Auto}&+ {\scriptsize  Box} & {\scriptsize + Scribble} &  {\scriptsize  Auto} & {\scriptsize + Box} &{\scriptsize + Scribble}\\
\end{tabular}
\caption{\footnotesize Qualitative results on DAVIS2017 val. \textbf{Auto}: SiamMask, \textbf{Box}: Results after box correct. (Curve-VOT), \textbf{Scribble}: Results after Scribble-VOS.}
\label{fig:compare_DAVIS}
\end{figure*}

\begin{figure*}[t!]
\addtolength{\tabcolsep}{-1pt}
\begin{tabular}{cccc}
\includegraphics[width=0.248\linewidth, trim=500 400 500 200,clip]{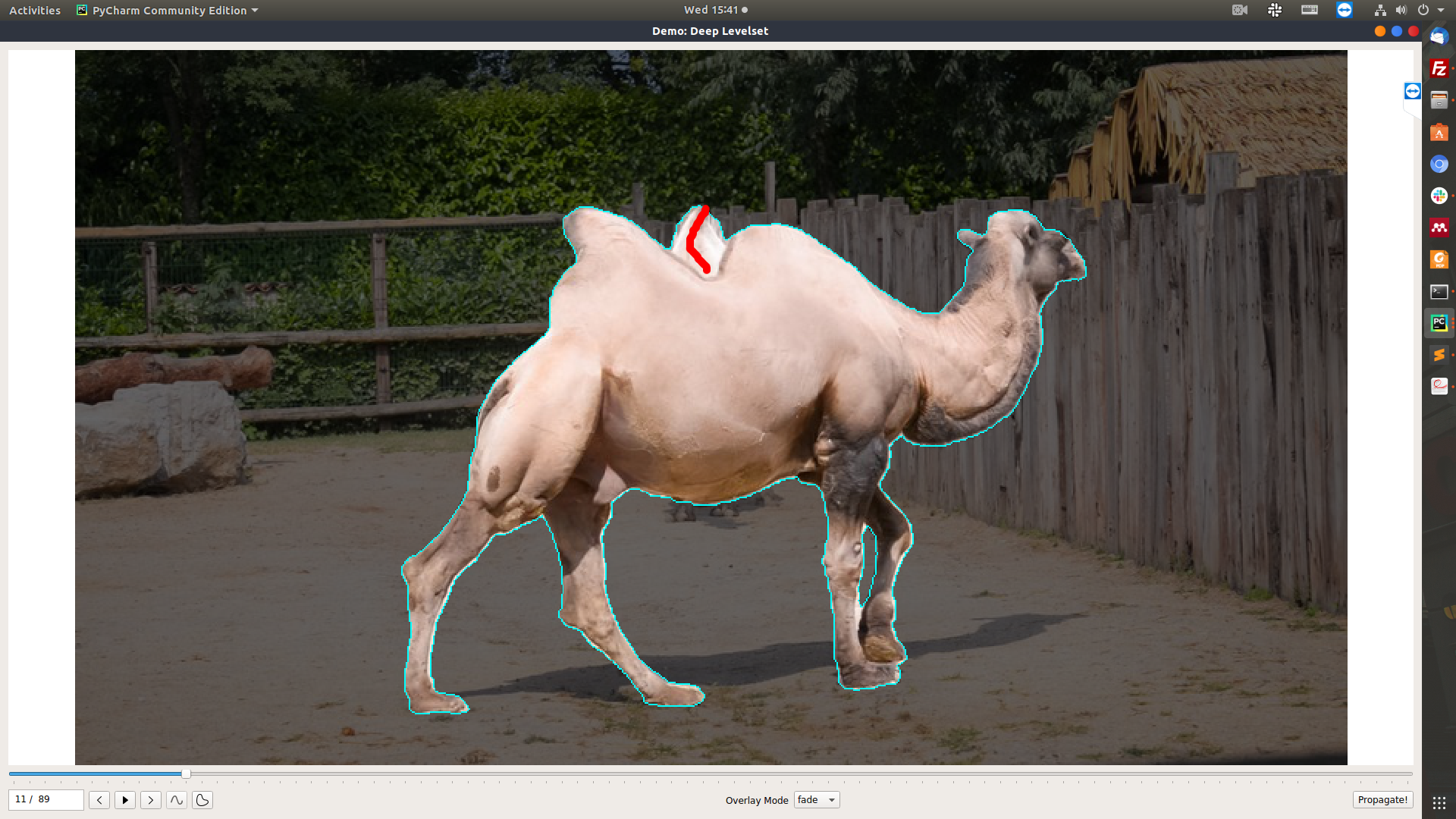} &  
\includegraphics[width=0.248\linewidth, trim=500 400 500 200, clip]{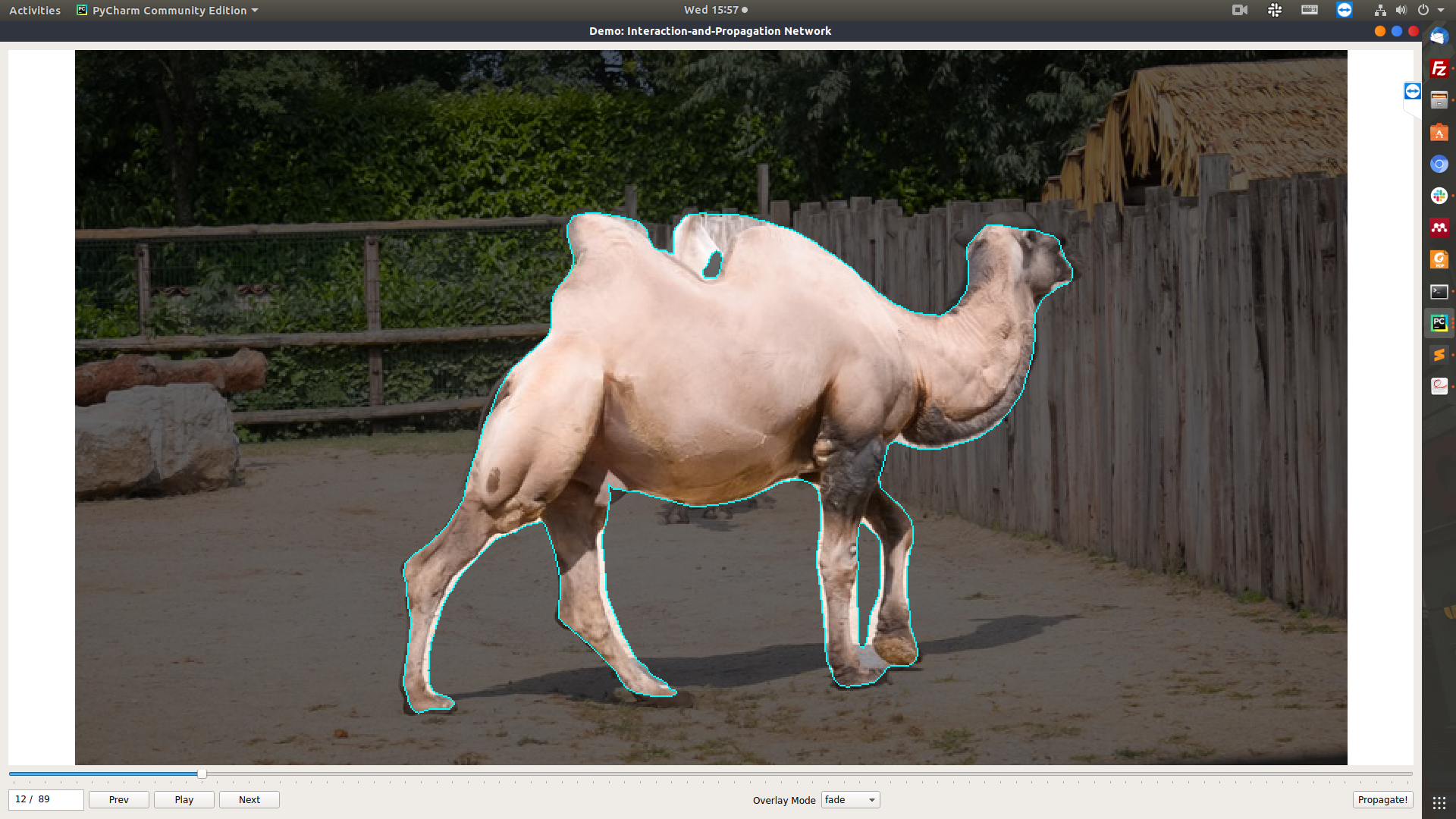}  & 
\includegraphics[width=0.248\linewidth, trim=500 400 500 200,clip]{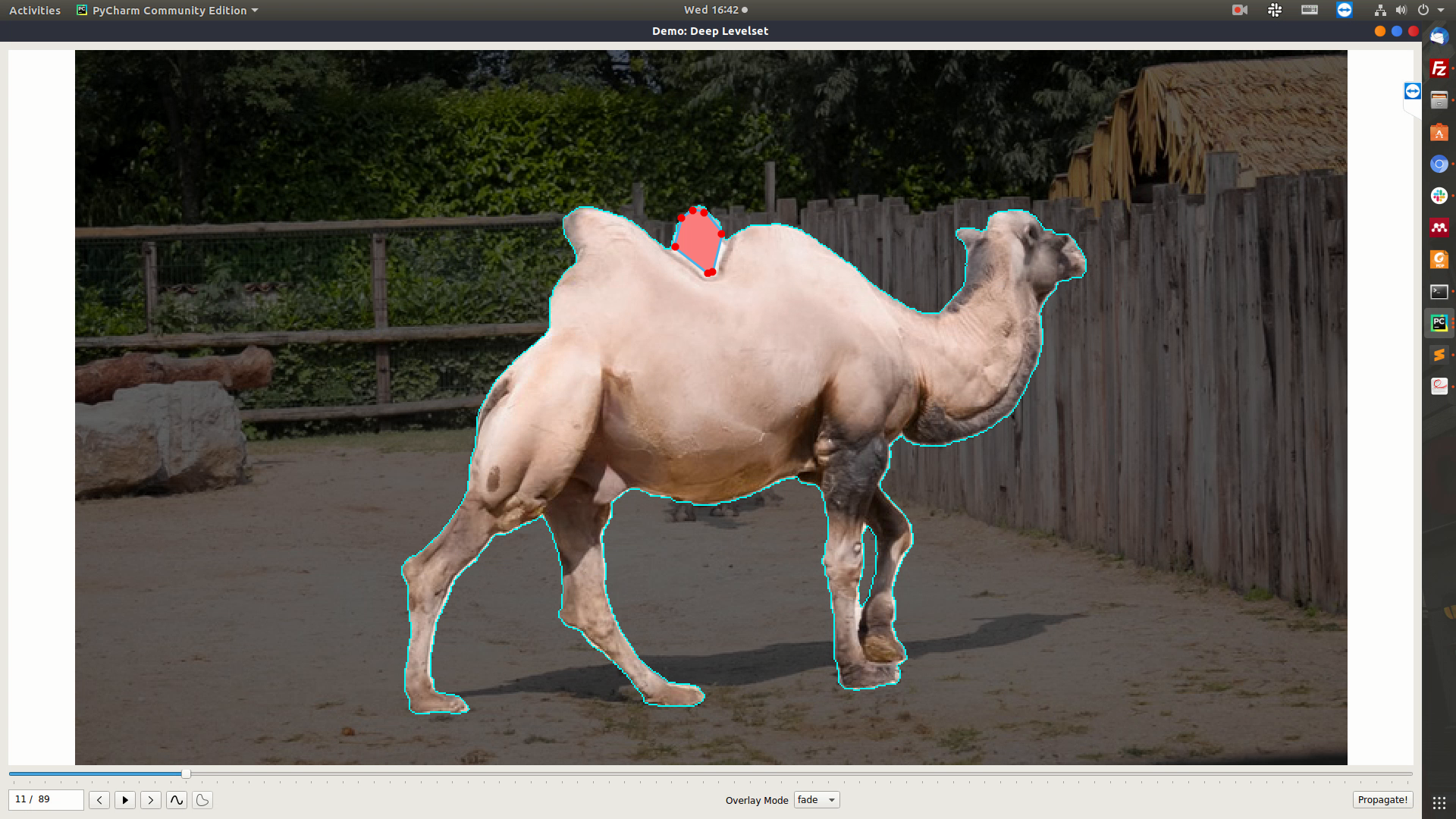} &  
\includegraphics[width=0.248\linewidth, trim=500 400 500 200, clip]{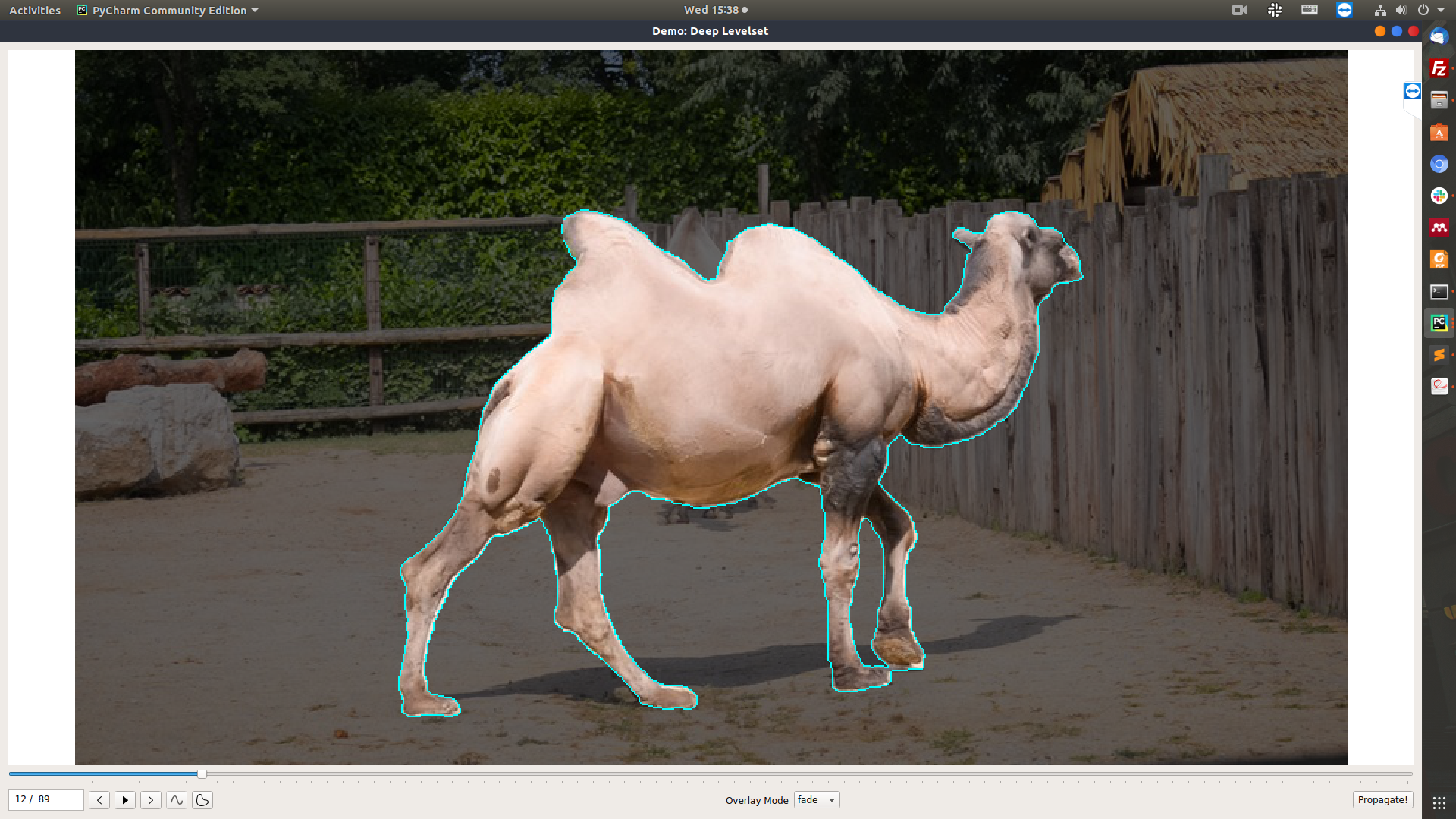}  \\ [-1mm]
{\scriptsize a) IPN correction at $t$} & {\scriptsize b) IPN at $t+1$} &  {\scriptsize  c) Our correction at $t$} & {\scriptsize  d) Ours at $t+1$}\\
\end{tabular}
\caption{\footnotesize Qualitative example demonstrating \textbf{effectiveness of our scribble propagation}. a) \& c): user draws a scribble in frame $t$. Note that IPN uses skeleton-like scribbles while we use trace scribbles. b) \& d): results after scribble propagation at frame t+1.}
\label{fig:ipn_ours_comparison}
\end{figure*}

\textbf{Qualitative Results:} We show qualitative results on DAVIS2017 val in 
Fig.~\ref{fig:results_DAVIS}. All results shown are annotated via Curve-VOT (9.14 clicks on average) following 5 rounds of scribble correction. Qualitative examples are in Fig.~\ref{fig:compare_DAVIS}. Results indicate that Curve-VOT corrects large errors and Scribble-VOS further refines them. Fig.~\ref{fig:ipn_ours_comparison} shows an example indicating the importance of our scribble propagation and differences between Thin-Line-Scribble and our Area-Scibble. 

\subsection{Out-of-Domain Annotation}
\begin{figure*}[t!]
\addtolength{\tabcolsep}{-0.8pt}
\begin{tabular}{ccc}
\includegraphics[height=1.3cm, trim=30 12 50 1,clip]{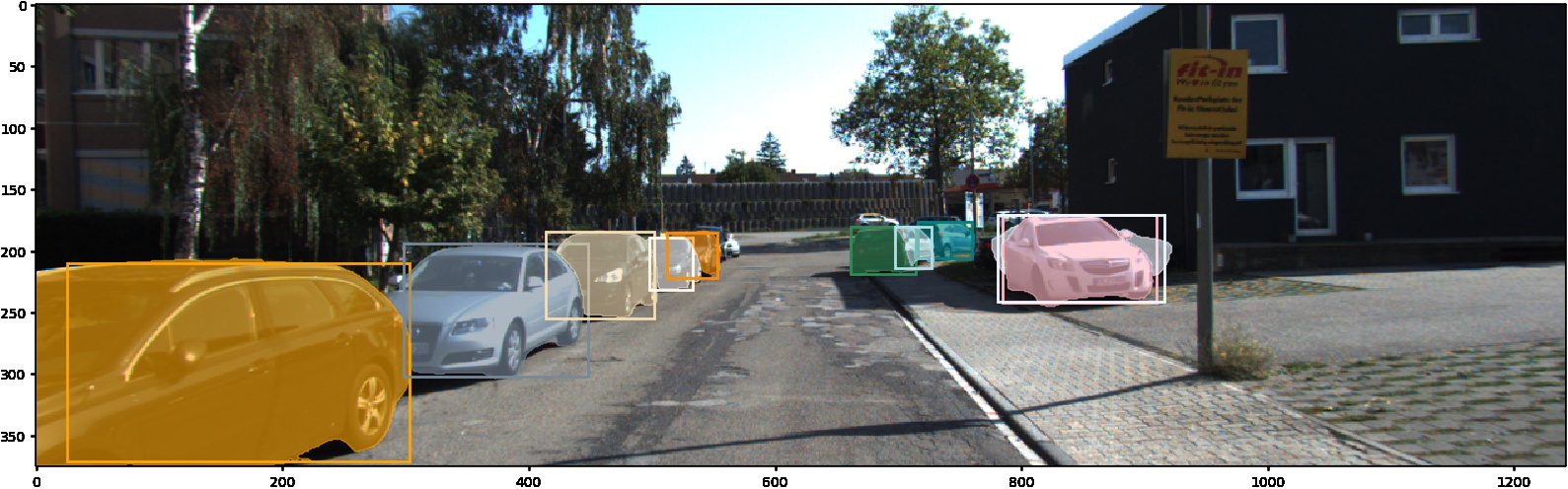} &
\includegraphics[height=1.3cm, trim=30 12 1 1,clip]{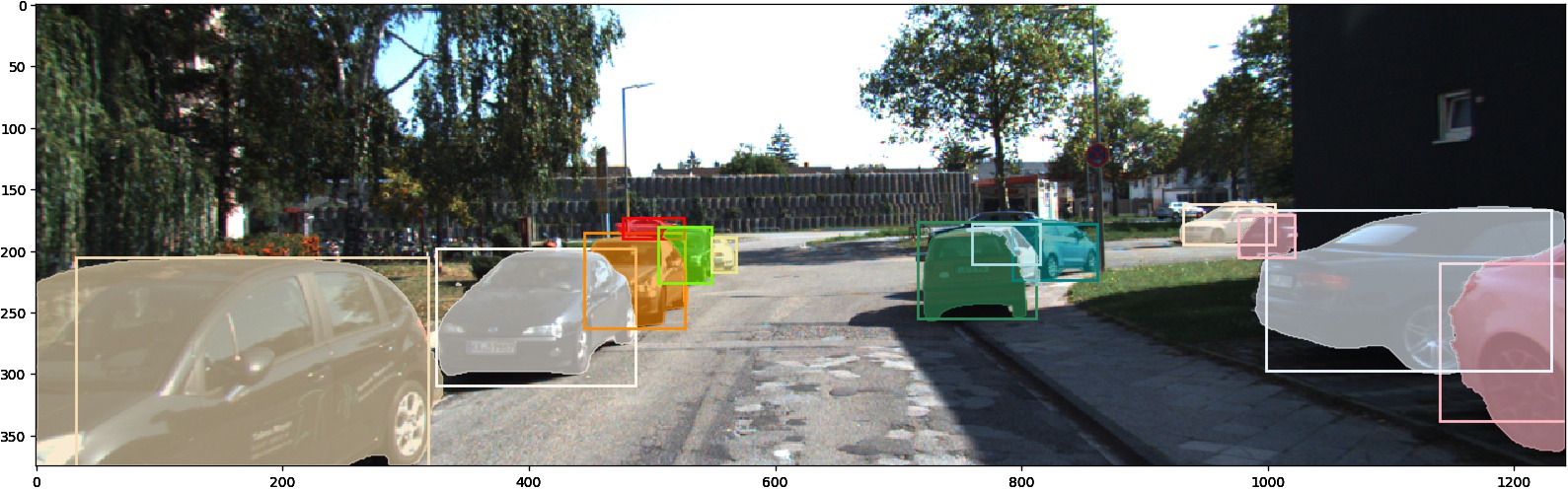} &
\includegraphics[height=1.3cm, trim=30 12 50 1, clip]{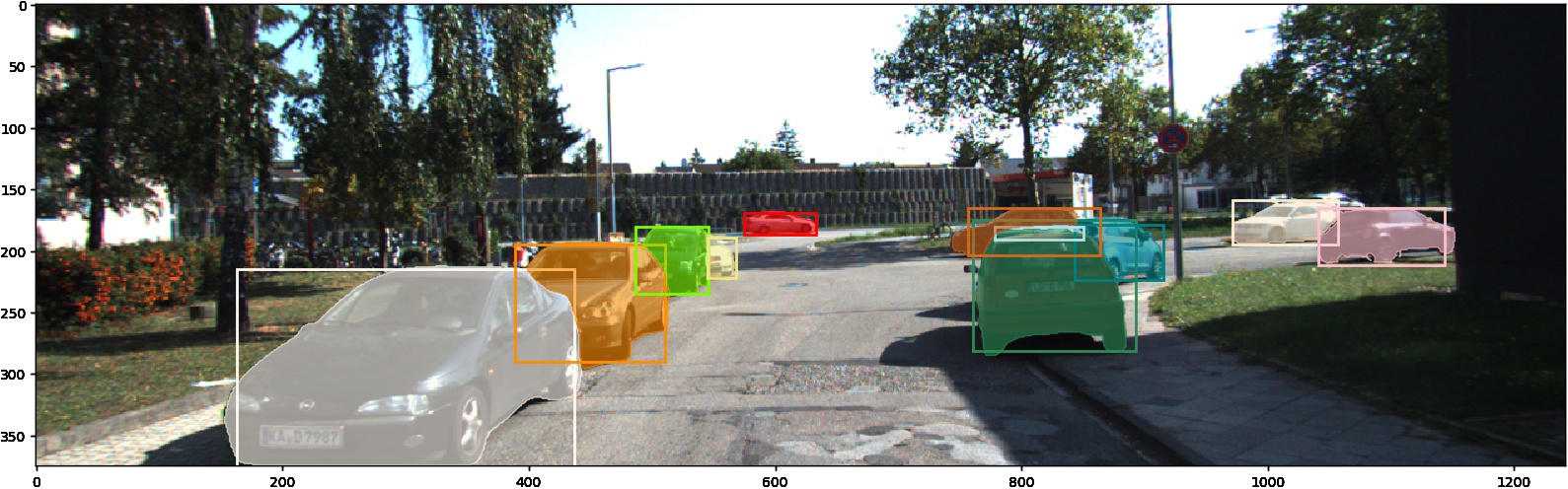} \\[-1mm]
\includegraphics[height=1.3cm, trim=30 12 50 1,clip]{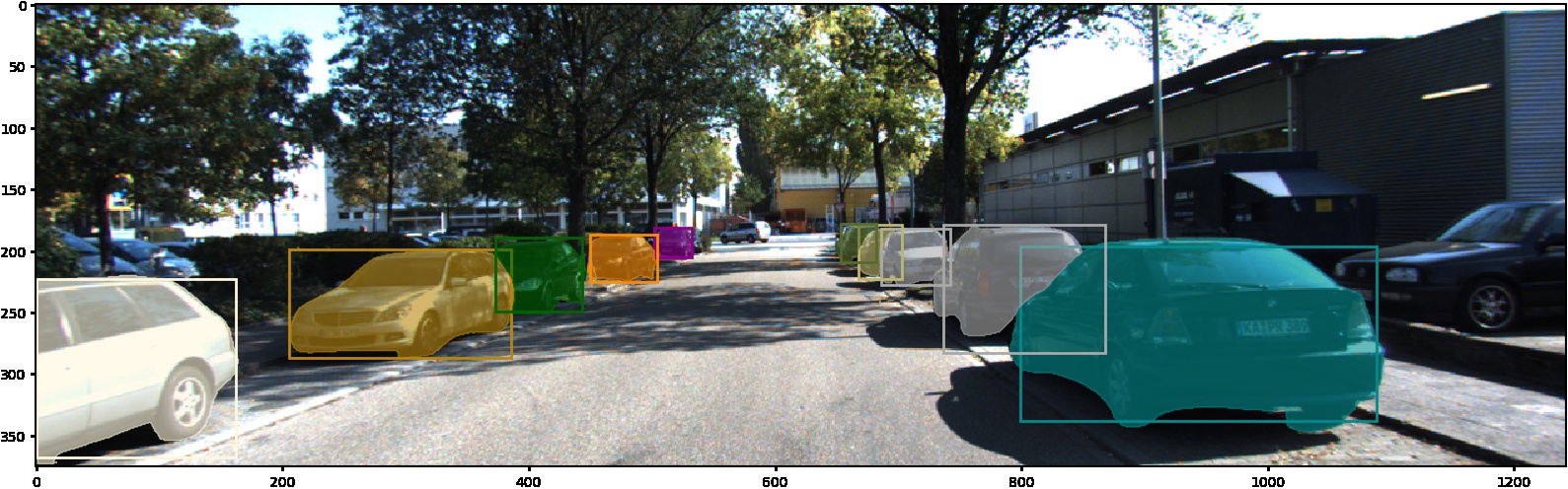} &
\includegraphics[height=1.3cm, trim=30 12 1 1,clip]{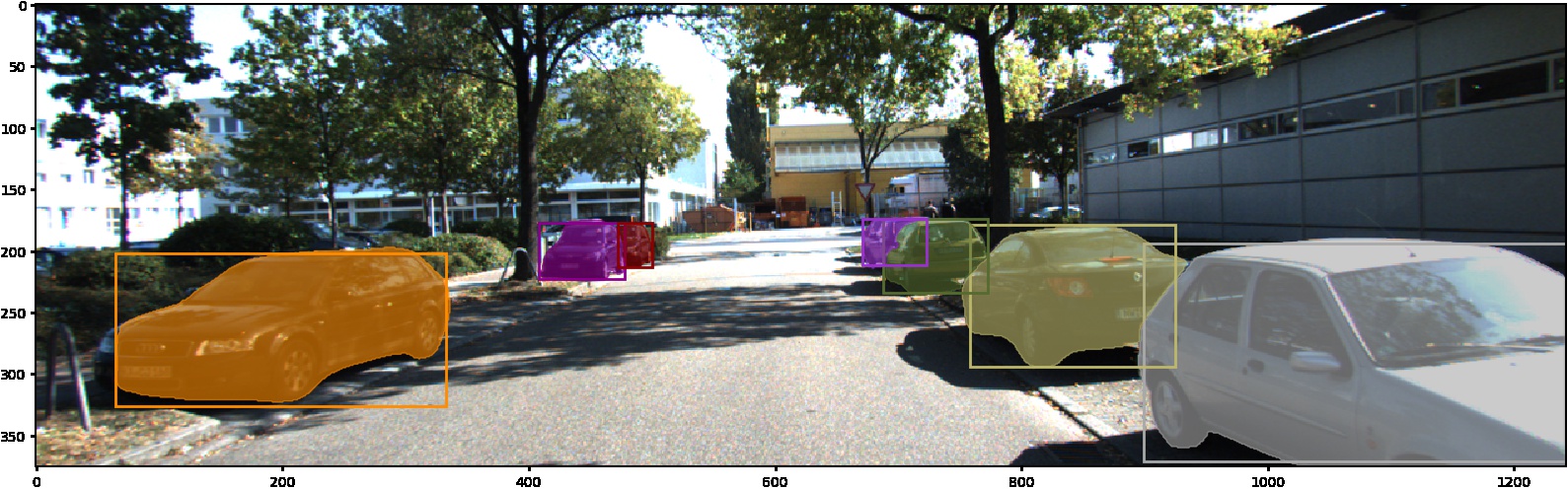} &
\includegraphics[height=1.3cm, trim=30 12 50 1, clip]{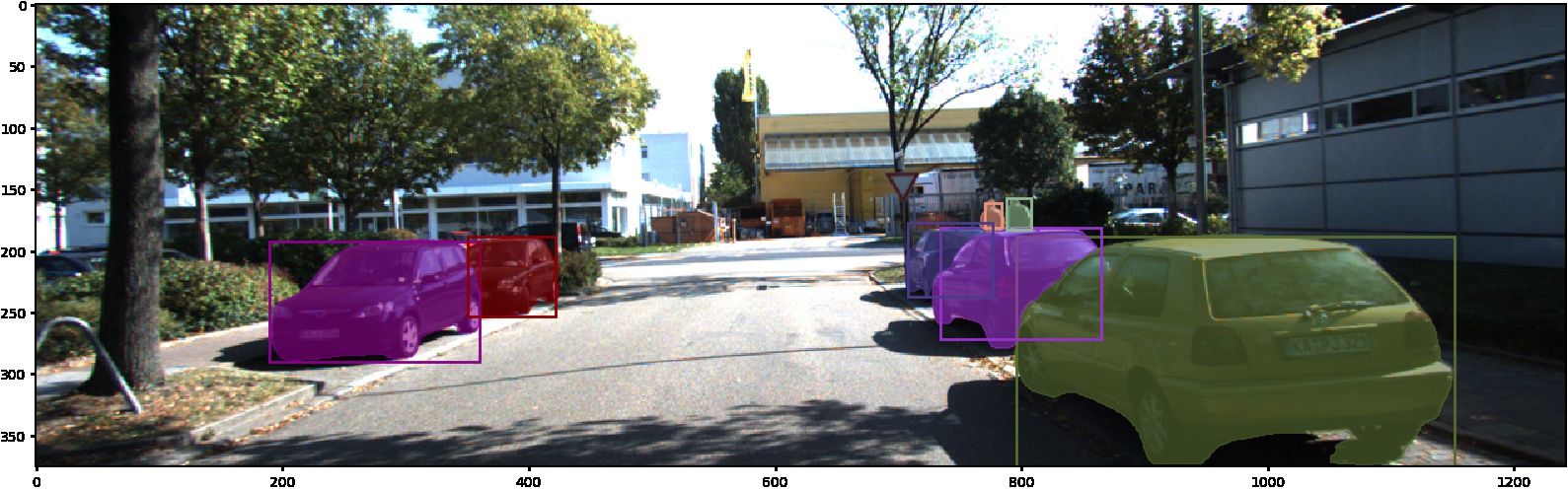} \\
\end{tabular}
\caption{\footnotesize Qualitative results on MOTS-KITTI~\cite{voigtlaender2019mots} using our full annotation framework. 
 5 rounds of scribble correction + 11.2 box corrections were used.}
\label{fig:results_KITTI}
\end{figure*}

\begin{figure*}[t!]
\addtolength{\tabcolsep}{-0pt}
\begin{tabular}{ccc}
\includegraphics[height=1.65cm, trim=680 150 275 130,clip]{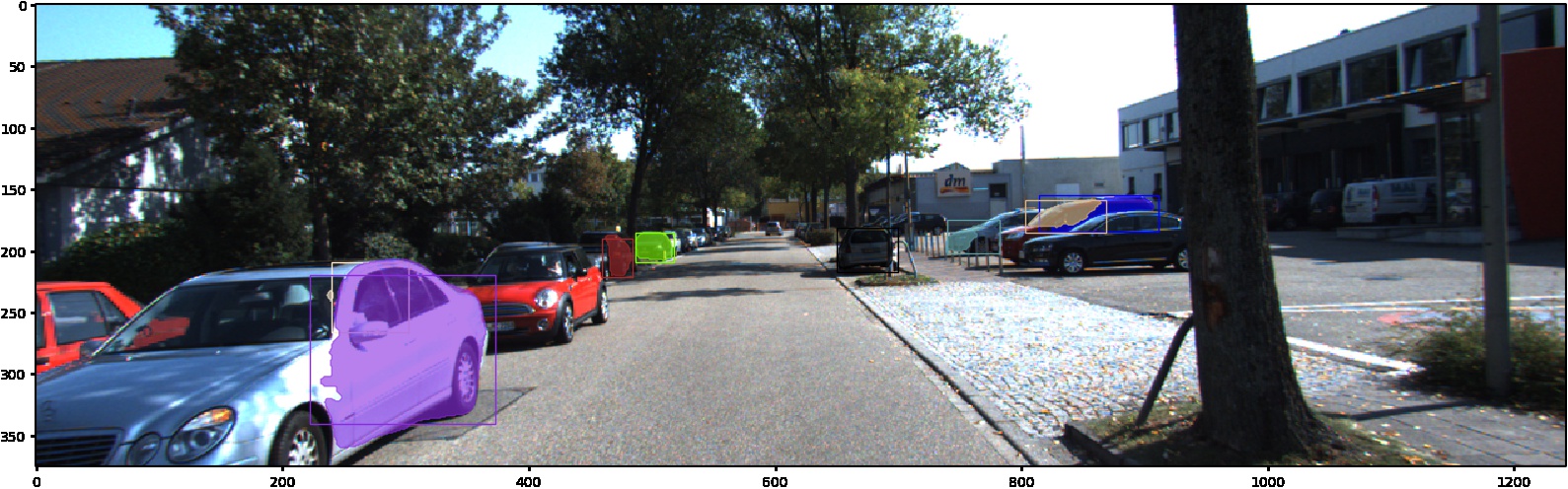} & 
\includegraphics[height=1.65cm, trim=680 150 275 130, clip]{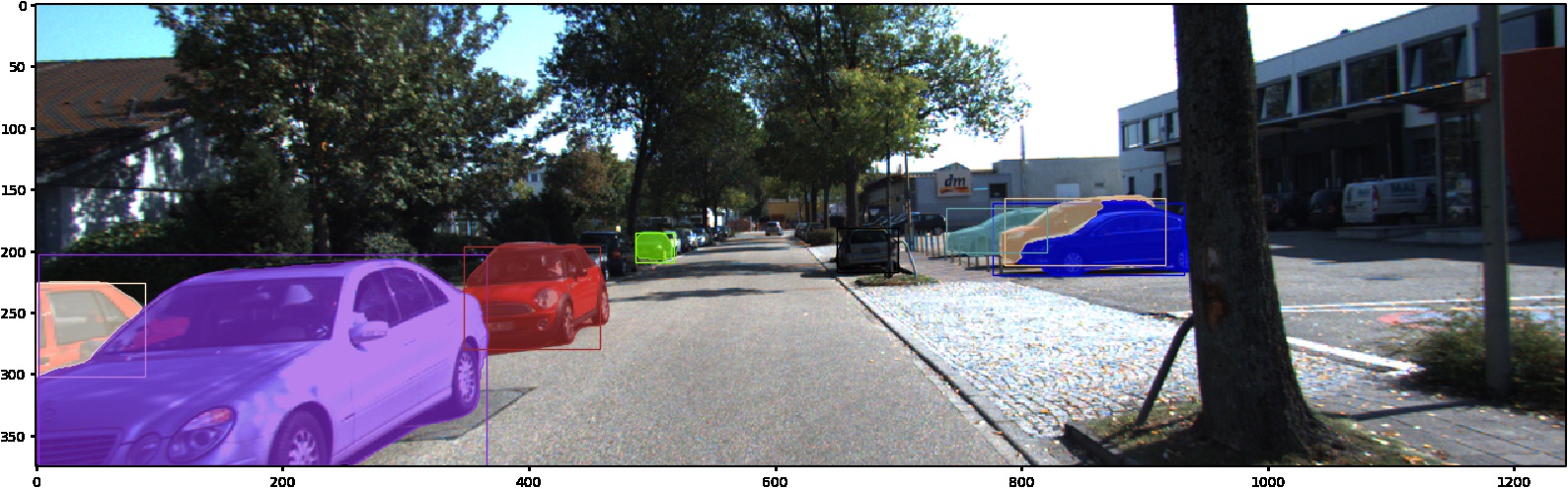}  & 
\includegraphics[height=1.65cm, trim=680 150 275 130, clip]{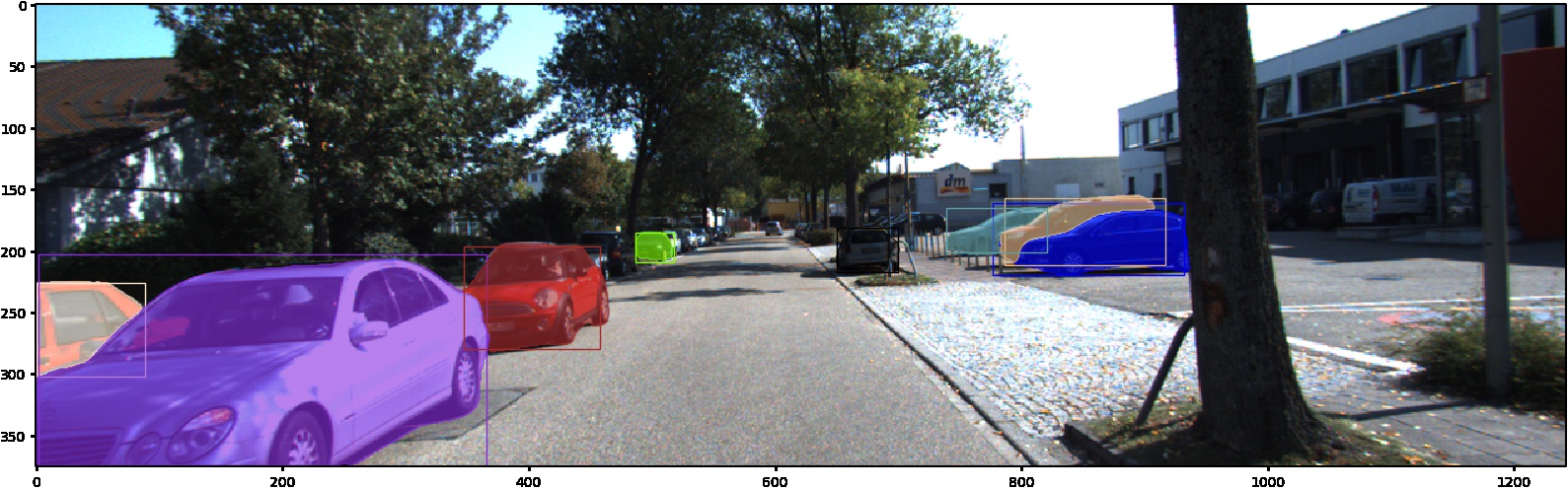} \\[-1.5mm]
{\scriptsize Auto}&+ {\scriptsize  Box} & {\scriptsize + Scribble}
\end{tabular}
\caption{\footnotesize Qualitative results on MOTS-KITTI~\cite{voigtlaender2019mots}. \textbf{Auto}: SiamMask, \textbf{Box}: Results after box correct. (Curve-VOT), \textbf{Scribble}: Results after Scribble-VOS.}
\label{fig:compare_KITTI}
\end{figure*}

We now run inference of our model on an unseen dataset (MOTS-KITTI). As shown in Fig.~\ref{fig:bbox_davis}(right), Curve-VOT outperforms the baseline by a large margin. Note that objects in KITTI are typically smaller than DAVIS, and so we adopt a smaller expansion size. Qualitative interactive tracking results are shown in Fig.~\ref{fig:results_KITTI} and Fig.~\ref{fig:compare_KITTI}. The car with blue mask in Fig.~\ref{fig:compare_KITTI} also shows our intuition about why we need two-stage annotation. Curve-VOT first ensures that the intended annotated object is not lost. This usually happens when multiple small and similar objects are adjacent.  
\subsection{User Study}
We put our framework to practice and show annotation results using human annotators with a simple tool that runs our models in the backend. All human experiments were done on the same desktop with a Nvidia-Titan Xp GPU. 



\textbf{In-Domain:}  We conduct a user study on the DAVIS2017 validation set. We randomly select one object per video which adds up to 1999 frames including blank frames. We employed four in-house annotators. Each was asked to annotate the same videos using both the baseline (IPN) and our method. For our method, we evaluate two models trained and tested at different resolutions: 256 (same as IPN) and 512, and we refer to them as ScibbleBox-256, ScibbleBox-512, respectively. For fairness, annotators first annotate with our model so that they gain familiarity with the data before using the baseline method. We ask annotators to annotate until there is no visible improvement. We show the mean curve for J\&F v.s annotation time in seconds in Fig.~\ref{fig:userstudy}. We include the data processing and model running times in the cumulative annotation time. We use filled area to denote variance between annotators. Note that the starting point of our model is to the right of our baseline because we only calculate J\&F after Curve-VOT corrections and their corresponding J\&F have already outperformed baseline's converging performance by a large margin. We also calculate mean J\&F in the overlapping time interval, which gives IPN: 79.25\%, ScribbleBox-256: 83.62\% and  ScribbleBox-512: 85.66\%.

We further report model running times in Table~\ref{tbl:running_time}.  We report  times with feature cache as ``Mask Prop. w/Cache''. Although our model with 512-resolution is slower, it outperforms both 256-resolution model and baseline given the same annotation time budget.

\textbf{Annotating new datasets:} We annotate a subset of the EPIC-Kitchen~\cite{Damen2018EPICKITCHENS,epic2020} dataset using our tool.  As shown in Fig.~\ref{fig:epic}, without any finetuning on out of domain data, our method generates high-quality masks with only a few annotations. 
We plan to annotate and make available a large portion of the dataset.

\begin{figure*}[t!]
\begin{tabular}{ccccc}
\includegraphics[width=0.195\linewidth, trim=0 0 0 0,clip]{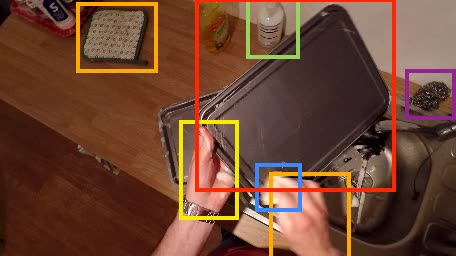} &
\includegraphics[width=0.195\linewidth, trim=0 0 0 0,clip]{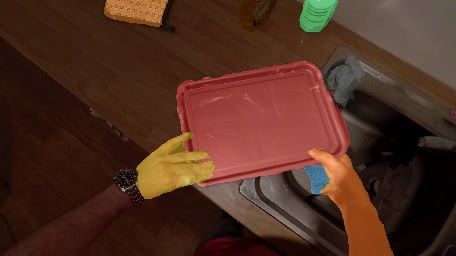} &
\includegraphics[width=0.195\linewidth, trim=0 0 0 0,clip]{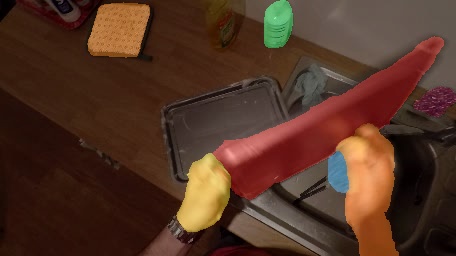} &
\includegraphics[width=0.195\linewidth, trim=0 0 0 0,clip]{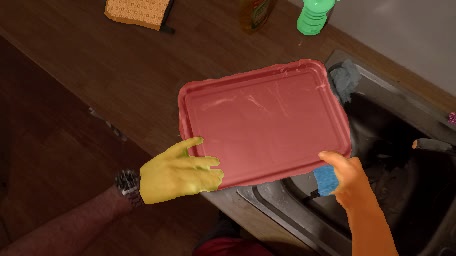} &
\includegraphics[width=0.195\linewidth, trim=0 0 0 0,clip]{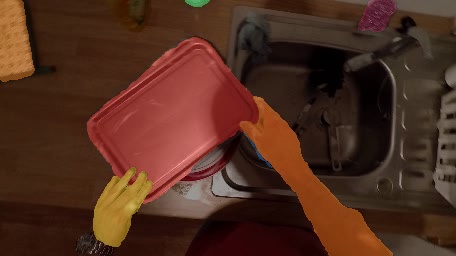} \\[-0.5mm]

\includegraphics[width=0.195\linewidth, trim=0 0 0 0,clip]{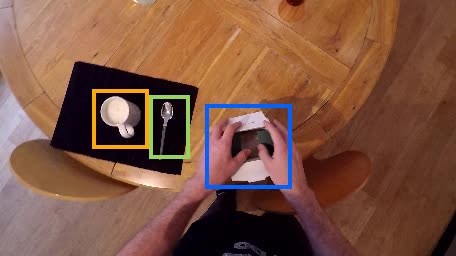} &
\includegraphics[width=0.195\linewidth, trim=0 0 0 0,clip]{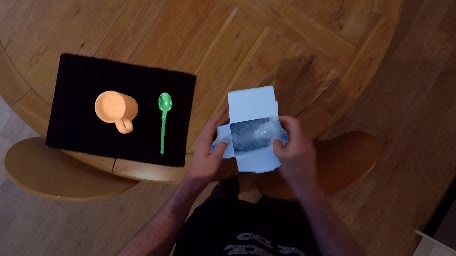} &
\includegraphics[width=0.195\linewidth, trim=0 0 0 0,clip]{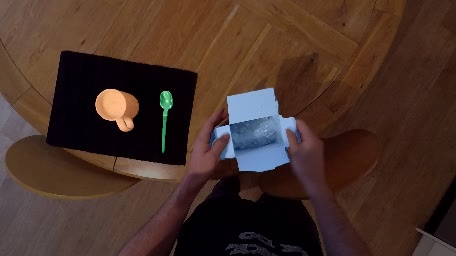} &
\includegraphics[width=0.195\linewidth, trim=0 0 0 0,clip]{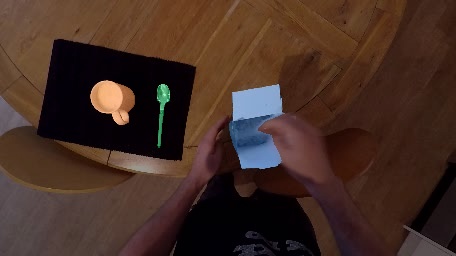} &
\includegraphics[width=0.195\linewidth, trim=0 0 0 0,clip]{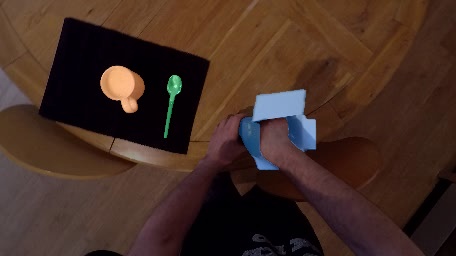} 
\end{tabular}
\caption{\footnotesize Example annotations using our annotation tool on the EPIC-Kitchen dataset. Each object in a 100-frame video requiring on average 69.87s of annotation time (including inference time). The first column indicates target objects.}
\label{fig:epic}
\end{figure*}

\vspace{-5mm}
\section{Conclusion}
\vspace{-5mm}
\label{sec:conc}

We introduced a novel video annotation framework that splits annotation into two steps: annotating box tracks, and labeling masks inside these tracks. Box tracks are annotated efficiently by approximating the trajectory using a parametric curve with a small number of control points which the annotator can interactively correct. Segmentation masks are corrected via scribbles which are efficiently propagated through time. We showed significant performance gains in annotation efficiency over past work in two major benchmarks. 
\vspace{-3mm}
\begin{flushleft}
\footnotesize{\textbf{Acknowledgments.} This work was supported by NSERC. SF acknowledges the Canada CIFAR AI Chair award at the Vector Institute.}
\end{flushleft}

{\small
\bibliographystyle{ieee}
\bibliography{egbib}
}
\newpage
\section{Appendix}
In the supplementary material, we provide a more detailed description of interactive segmentation training and the mask decoder's structure. We also show additional qualitative results, and provide an overview of our annotation tool when used in practice.

\vspace{-5mm}
\subsection{Interactive Segmentation Training Details}
\vspace{-3mm}

\textbf{Multi-Stage Training For Mask Propagation Module:}
The \emph{mask propagation module} is trained in two stages. We pretrain the module on synthetic video clips. Each clip includes a pair of reference and target frame, and is generated by applying random affine transformation and object composition following ~\cite{wug2018fast}. We expect this to make the network more robust to variations in object appearance. We then fine-tune it on video segmentation datasets. For each training video clip, we randomly sample 3 ordered frames and apply data augmentation including random flipping, $10\%$ bounding box noise and affine transformation. The two-stage training takes about 1 day on synthetic video clips, 3 days on DAVIS2017 and 5 days on YoutubeVOS using 4 NVIDIA Tesla P100 GPUs. 

\textbf{Batch Hard Triplet Loss For Scribble Propagation Module:} We utilize the batch hard triplet loss from~\cite{Chen_2018} to supervise the embedding head in the scribble propagation network. 
First, we use $f_c\in \mathbb{R}^D$ to denote the feature vector at each location in the feature map $F_c$. 
Similarly, for each location in the image feature $F_r$ of frame $r$, we have a feature vector $f_r$. 
For each $f_r \in F_r$, if it corresponds to $f_c$, we define it as positive/true sample with respect to $f_c$, denoted by $f_r^+$, otherwise it is negative/false sample, denoted as $f_r^-$. We denote the set of $f_r^+$ as $\{f_r^+\}$ and the set of $f_r^-$ as $\{f_r^-\}$. Then the batch hard triplet loss can be written as: 
\begin{equation}
L_{\mathrm{BHTriplet}}(F_c, F_r) =  \sum_{f_c \in F_c} l ({f_c, \{f_r^+\}, \{f_r^-\}}),  
\end{equation}
where  
\begin{equation}
\begin{aligned}
l(f_c,\{f_r^+\}, \{f_r^-\}) = \min_{f_r^+ \in \{f_r^+\} }\lVert emb(f_c)-emb(f_r^+) \rVert_2^2- \\ \min_{f_r^-\in  \{f_r^-\} }\lVert emb(f_c)-emb(f_r^-) \rVert_2^2+\alpha
 \end{aligned}
\end{equation}
Here, $\alpha$ is the minimal margin between positive and negative samples, and $emb(\cdot)$ is the embedding function. 
This loss is able to push two samples that do not correspond to each other away even if they have similar semantic information.

\textbf{Implementation Details:}
We always crop the input images, masks (and scribbles) based on the input bounding box and resize them to $512\times 512$ (or $256 \times 256$ when specifing). 
We use SGD optimizer with momentum of 0.9 and a cyclical learning rate policy~\cite{Smith_2017} to speed up the training process. The minimum and maximum learning rate is set to 1e-5 and 1e-3, respectively. We use \textit{triangular2} CLR policy and 4 cycles throughout training.  Once mask propagation module is trained, we freeze the image encoder and train \emph{interactive segmentation module}  and \emph{scribble propagation module}.

\begin{figure*}[h!]
\vspace{-4mm}
\begin{tabular}{cc}
\includegraphics[width=0.48\linewidth,trim=0 250 0 0,clip]{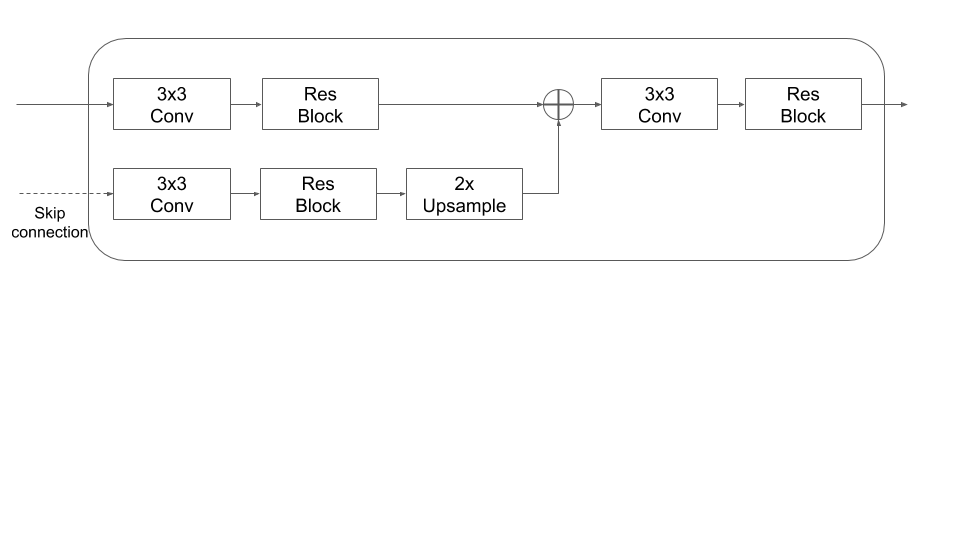}\hspace{0.5mm} & \hspace{0.5mm}
\includegraphics[width=0.48\linewidth, trim=0 300 150 0, clip]{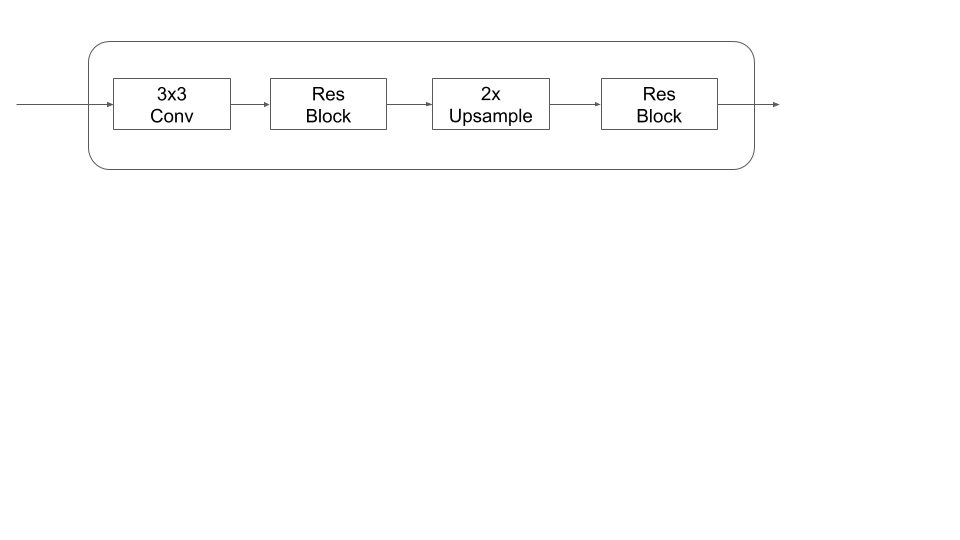}\hspace{0.5mm}\\
{\footnotesize Refinement layer 1 and 2} & {\footnotesize Refinement layer 3}
\end{tabular}
\caption{Structure of three refinement layers in the decoder. Note that the skip connection in the refinement layer 3 is removed.}
\label{fig:decoder}
\end{figure*}
\vspace{-7mm}
\begin{figure*}[t!]
\vspace{-7mm}
\addtolength{\tabcolsep}{-3.2pt}
\begin{minipage}{0.7\linewidth}
\begin{tabular}{cc}
\includegraphics[height=2.7cm, width=0.5\linewidth,clip]{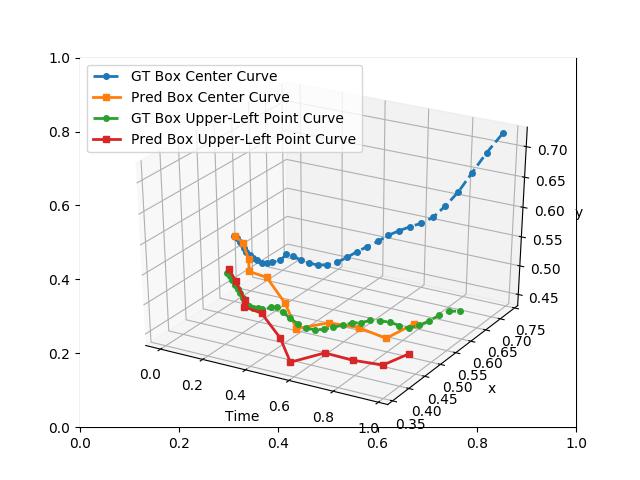} &
\includegraphics[height=2.7cm, width=0.5\linewidth,clip]{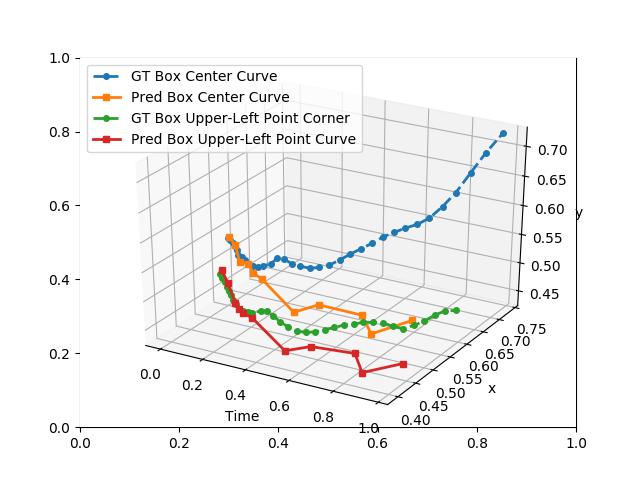} \\
{\small Auto}  & {\small 1 Corrections} \\
\includegraphics[height=2.7cm, width=0.5\linewidth,clip]{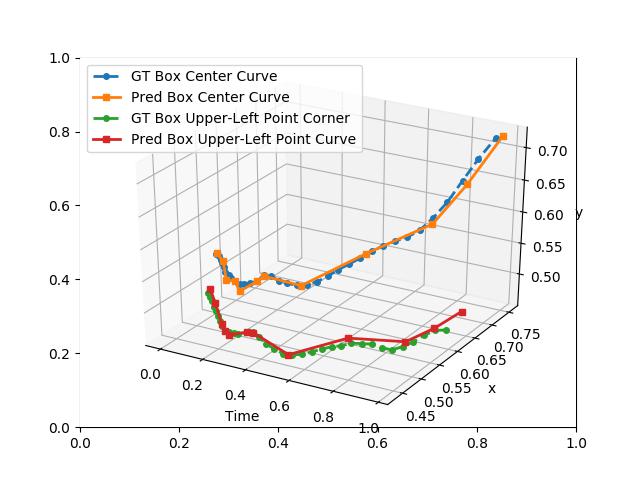} &
\includegraphics[height=2.7cm, width=0.5\linewidth,clip]{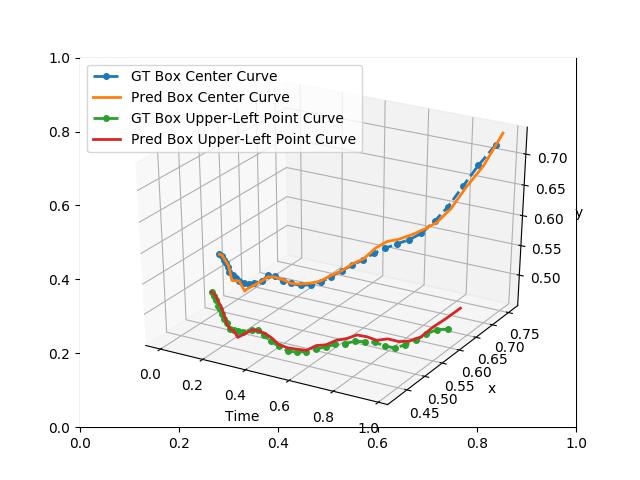} \\[-3mm]
{\small 2 Corrections} &{\small Box Refinement}\\
\end{tabular}
\end{minipage}
\begin{minipage}{0.27\linewidth}
\begin{center}
{
\vspace{4.2mm}
\caption{\footnotesize This plot shows an example of how a box track gets refined with user's corrections. Results are reported on KITTI. Blue \& green are ground-truth trajectories of the center and top-left box coordinate. Orange \& red is the track from Curve-VOT. Only 2 corrections are required to produce an accurate track.}
\label{fig:qualitative_human_tracking}
}
\end{center}
\end{minipage}
\vspace{-3mm}
\end{figure*}

\vspace{-5mm}
\subsection{Mask Decoder's Structure}
\vspace{-3mm}
There are two decoders in our framework.
\textit{Interactive decoder} is used to produce a refined mask of a single frame based on the human-in-the-loop interaction.  We also have a \textit{propagation decoder} which is  shared by both the mask and scribble propagation networks. 

In particular, we combine three refinement modules~\cite{wug2018fast} as our decoder. Different from the original structure, we remove the first refinement module and add an additional refinement module without skip connection before the last convolution layer. Three refinement modules produce feature maps with 224, 224, 128 channels, respectively, and the last convolution layer produces the final mask. The size of the output mask is half the size of the input image. 

The detailed structures of our refinement modules are shown in Figure~\ref{fig:decoder}.
\vspace{-5mm}
\subsection{Qualitative Results}
\begin{figure*}[hp!]

\begin{tabular}{ccc}
\includegraphics[width=0.33\linewidth,height=0.22\linewidth,trim=0 0 0 0,clip]{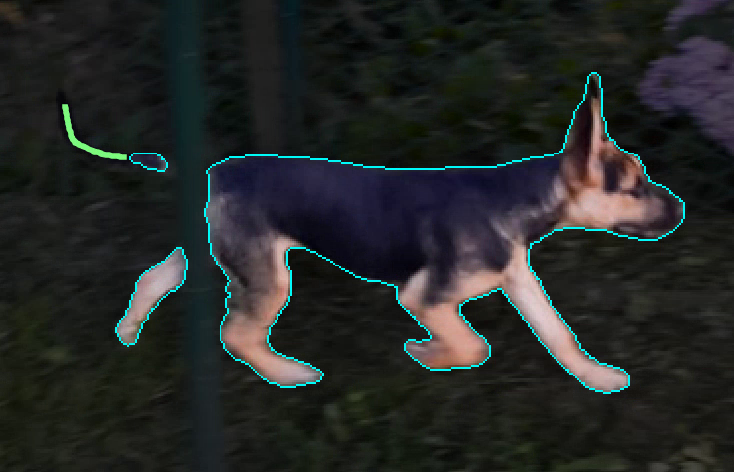}&
\includegraphics[width=0.33\linewidth,height=0.22\linewidth, trim=0 0 0 0, clip]{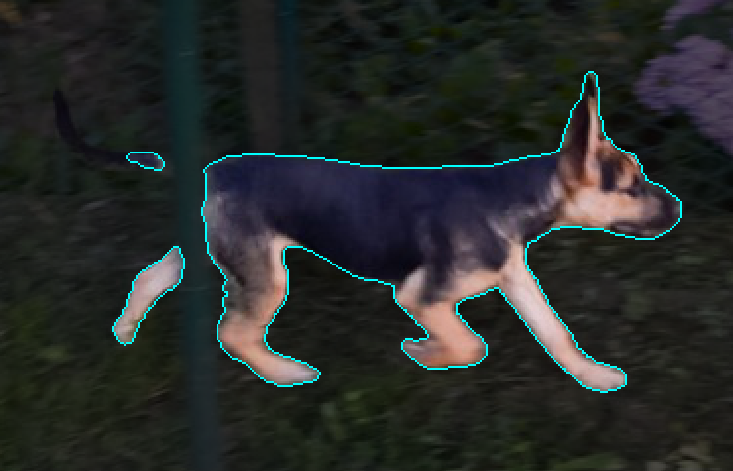}&
\includegraphics[width=0.33\linewidth, height=0.22\linewidth,trim=0 0 0 0, clip]{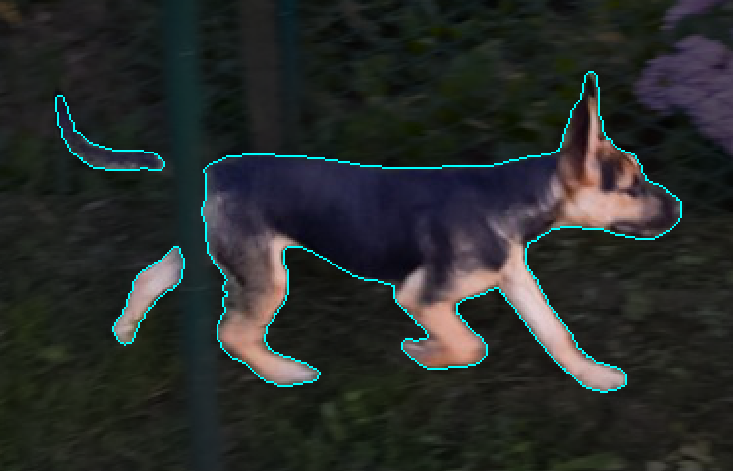}\\ 
\includegraphics[width=0.33\linewidth, height=0.22\linewidth,trim=30 20 0 40, clip]{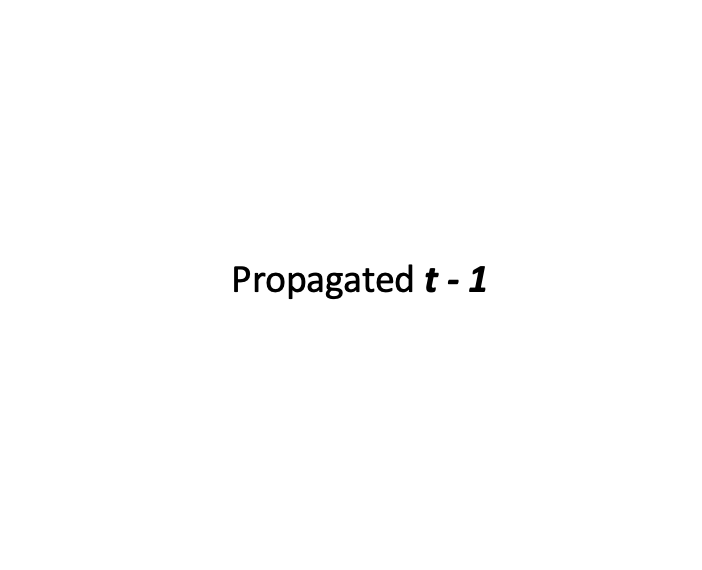}&
\includegraphics[width=0.33\linewidth, height=0.22\linewidth,trim=0 0 0 0,clip]{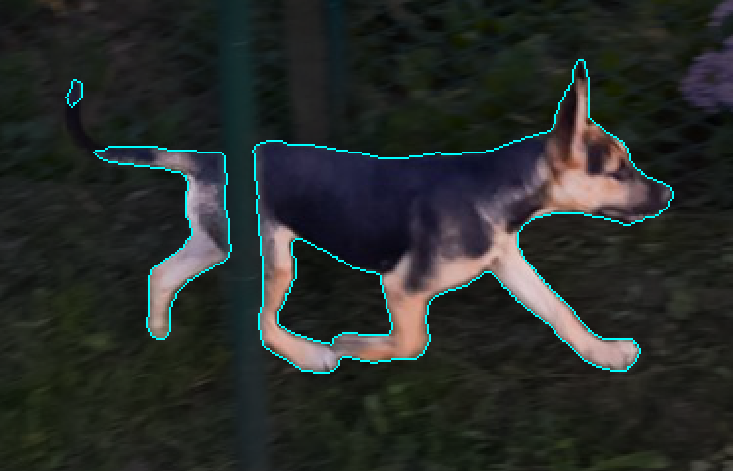}&
\includegraphics[width=0.33\linewidth,height=0.22\linewidth,trim=0 0 0 0,clip]{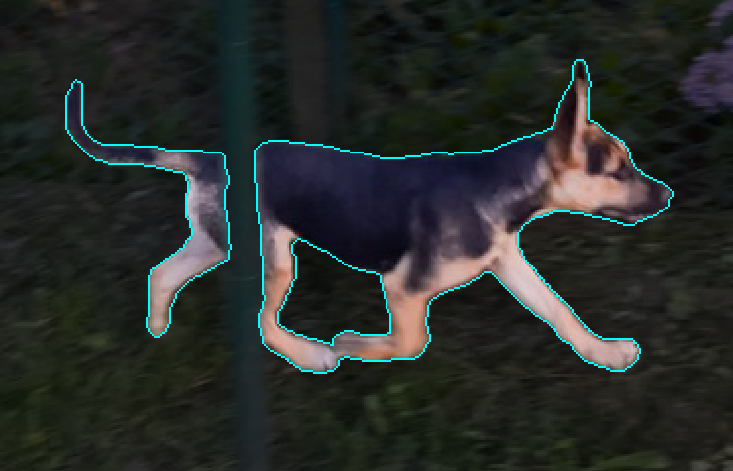}\\ 
\includegraphics[width=0.33\linewidth, height=0.22\linewidth,trim=10 0 0 0, clip]{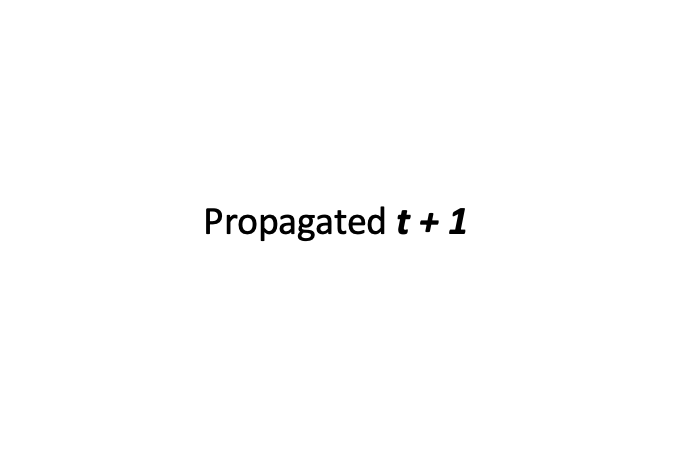}&
\includegraphics[width=0.33\linewidth,height=0.22\linewidth, trim=0 0 0 0, clip]{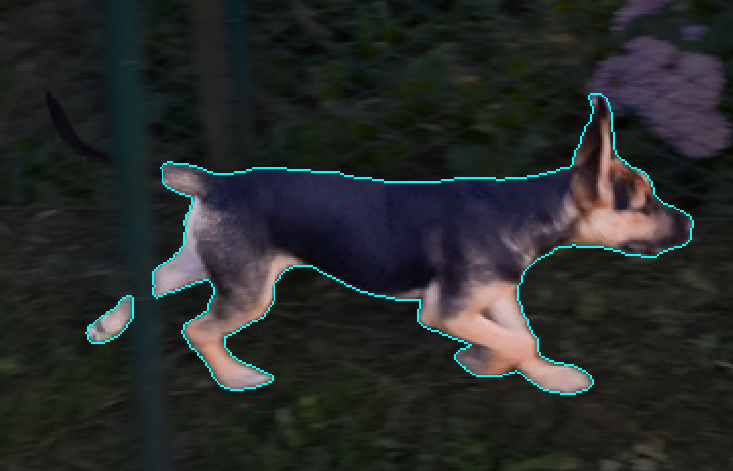}&
\includegraphics[width=0.33\linewidth,height=0.22\linewidth,trim=0 0 0 0,clip]{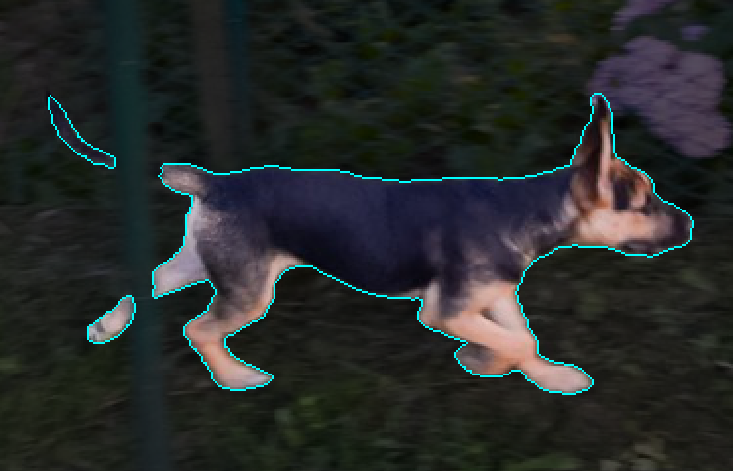}\\
 & {\footnotesize W/O Scribble Consistency} & {\footnotesize  W/ Scribble Consistency}\\
\end{tabular}
\vspace{-2mm}
\caption{Qualitative examples demonstrate the effectiveness of our scribble consistency loss. The second and third rows show neighbour frames propagated from $t$ with and without scribble consistency loss respectively.}
\label{fig:scl_comparison2}
\end{figure*}

\vspace{-3mm}
We illustrate an example Curve-VOT result in Fig.~\ref{fig:qualitative_human_tracking}.
In Figure~\ref{fig:scl_comparison2} we provide a qualitative comparison for our interactive segmentation network trained with and without the scribble consistency loss. The user input scribble is typically ignored without this loss.  

We show additional annotation examples on the EPIC-Kitchen dataset~\cite{Damen2018EPICKITCHENS} using our annotation tool in Figure~\ref{fig:epic}.

\subsection{Annotation Tool}
\begin{figure}[t!]
\vspace{-3mm}
\begin{minipage}{0.501\linewidth}
\begin{center}
\includegraphics[width=0.9\linewidth, trim=0 0 0 0,clip]{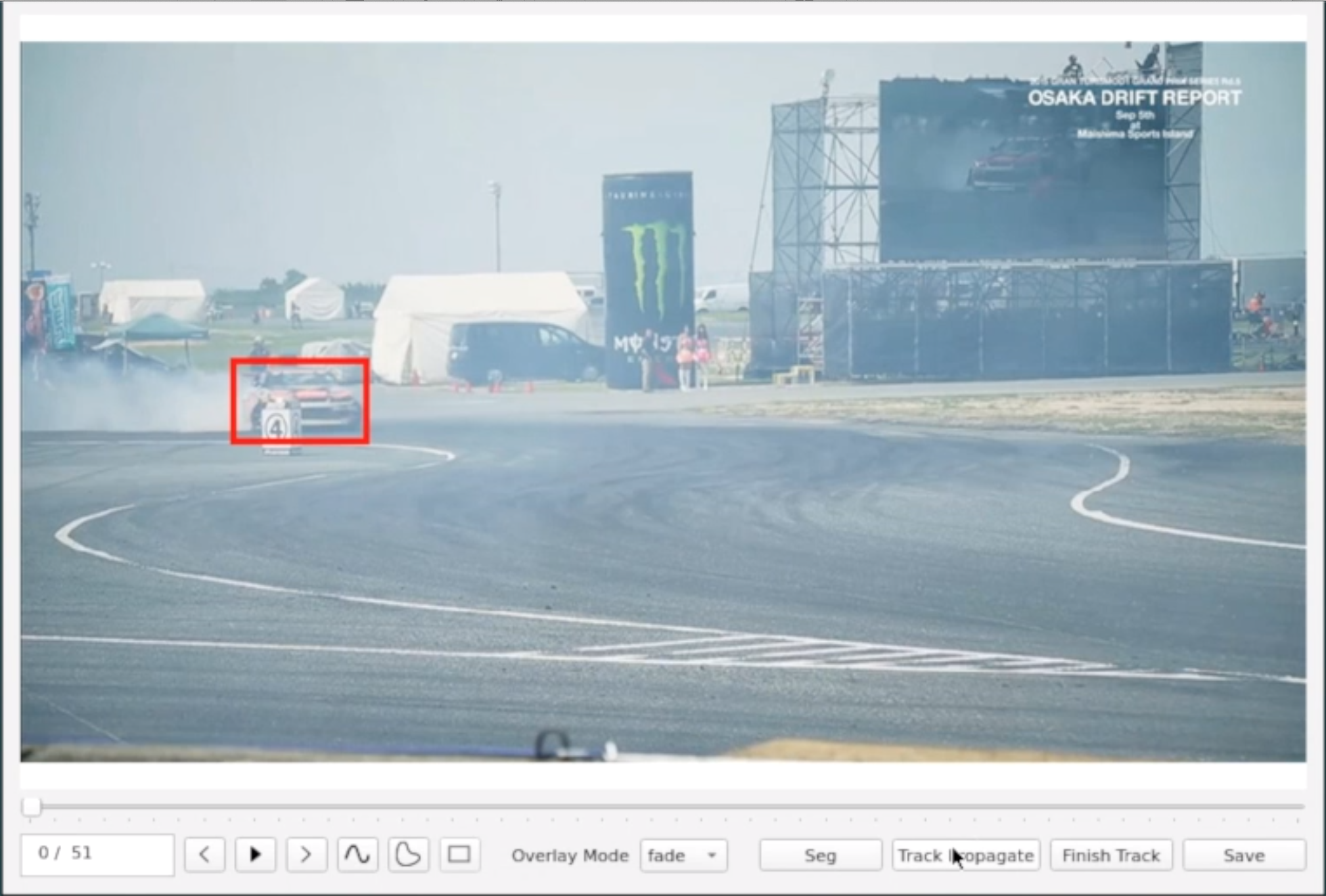}
\end{center}
\end{minipage}
\hspace{2mm}
\begin{minipage}{0.46\linewidth}
\vspace{3mm}
\textbf{Step1}: User draws a box at the first frame. 
\end{minipage}
\vspace{3mm}
\begin{minipage}{0.501\linewidth}
\begin{center}
\includegraphics[width=0.9\linewidth, trim=0 0 0 0,clip]{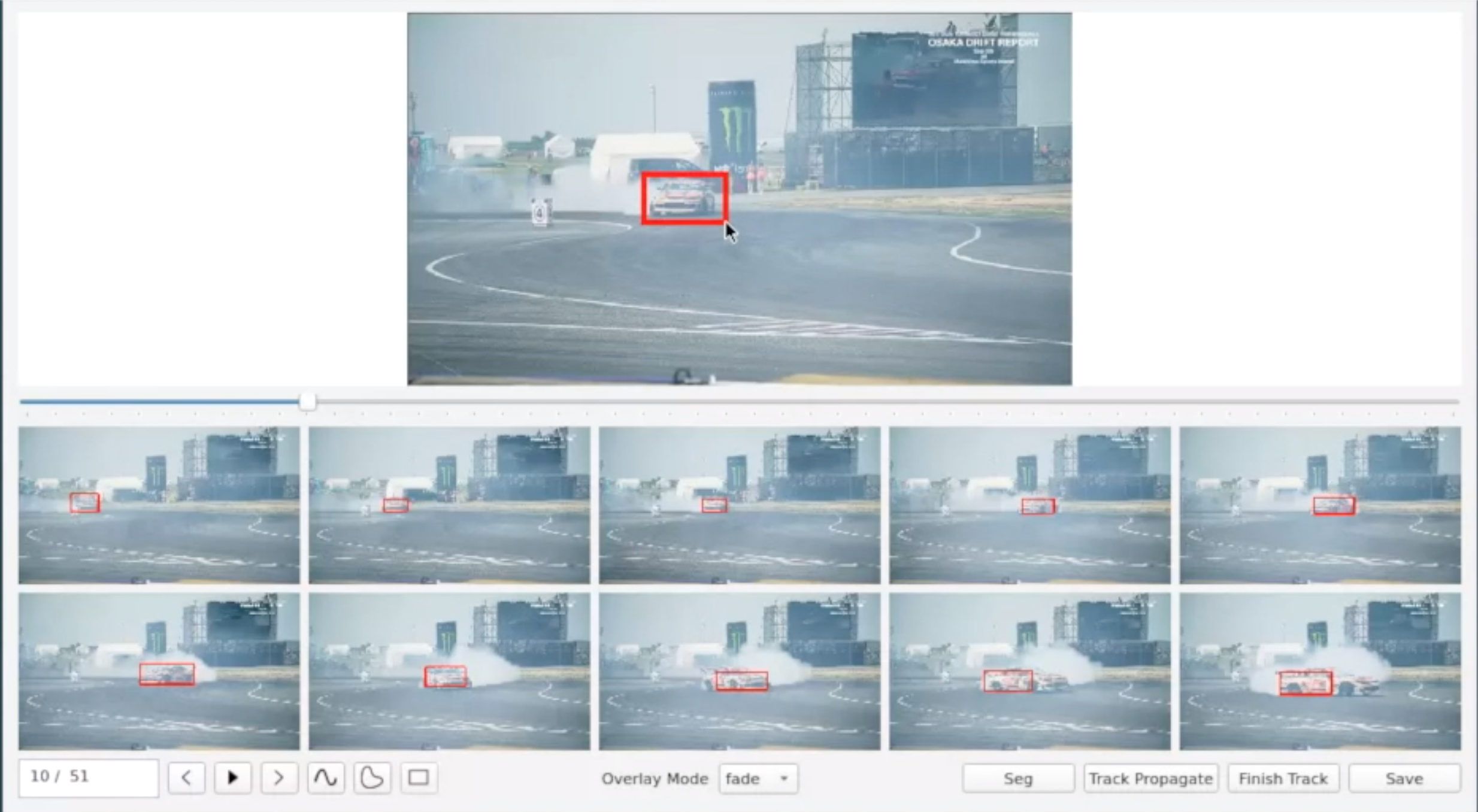}
\end{center}
\end{minipage}
\hspace{2mm}
\begin{minipage}{0.46\linewidth}
\vspace{3mm}
\textbf{Step2}: Auto tracking and pop out key frames (nearest frames to control points).
\end{minipage}

\vspace{3mm}
\begin{minipage}{0.501\linewidth}
\begin{center}
\includegraphics[width=0.9\linewidth, trim=0 0 0 0,clip]{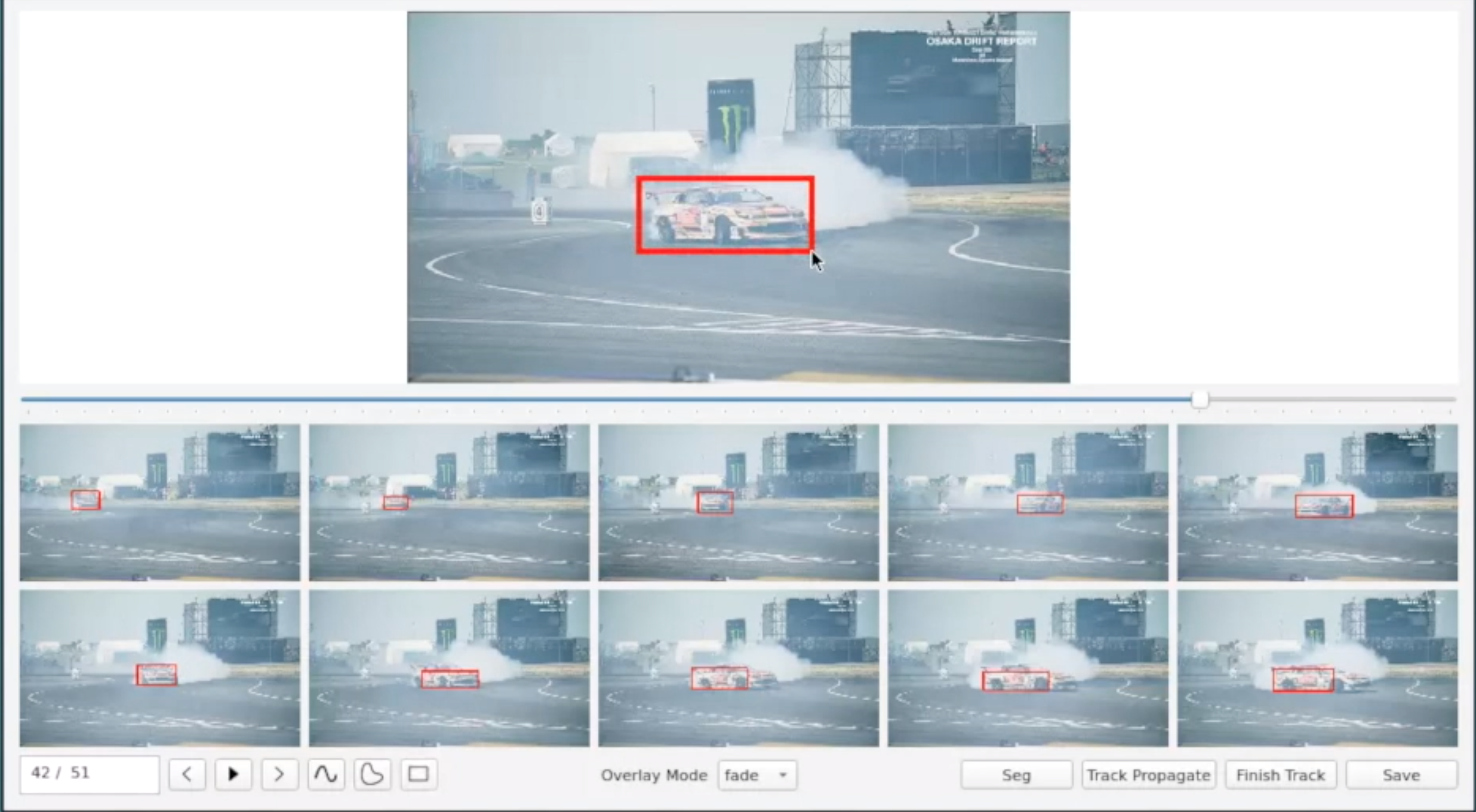}
\end{center}
\end{minipage}
\hspace{2mm}
\begin{minipage}{0.46\linewidth}
\vspace{3mm}
\textbf{Step3}: User spots a mistake in one of the keyframes and corrects it by drawing a new box. We then re-run the tracker and re-fit the curve.
\end{minipage}

\vspace{3mm}
\begin{minipage}{0.501\linewidth}
\begin{center}
\includegraphics[width=0.9\linewidth, trim=0 0 0 0,clip]{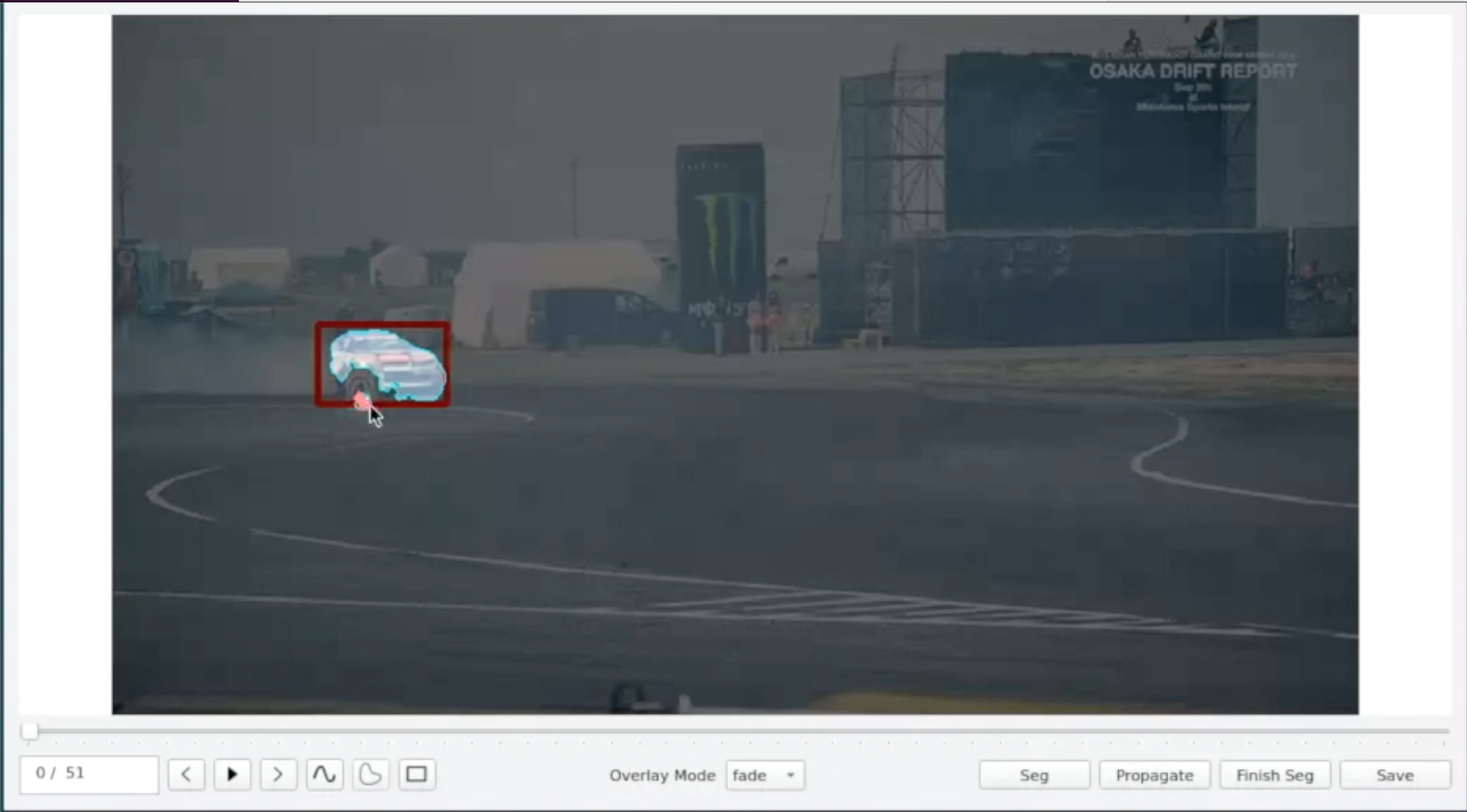}
\end{center}
\end{minipage}
\hspace{2mm}
\begin{minipage}{0.46\linewidth}
\vspace{3mm}
\textbf{Step4}: User corrects the mask with scribbles, which are propagated to other frames.
\end{minipage}

\vspace{3mm}
\begin{minipage}{0.501\linewidth}
\begin{center}
\includegraphics[width=0.9\linewidth, trim=0 0 0 0,clip]{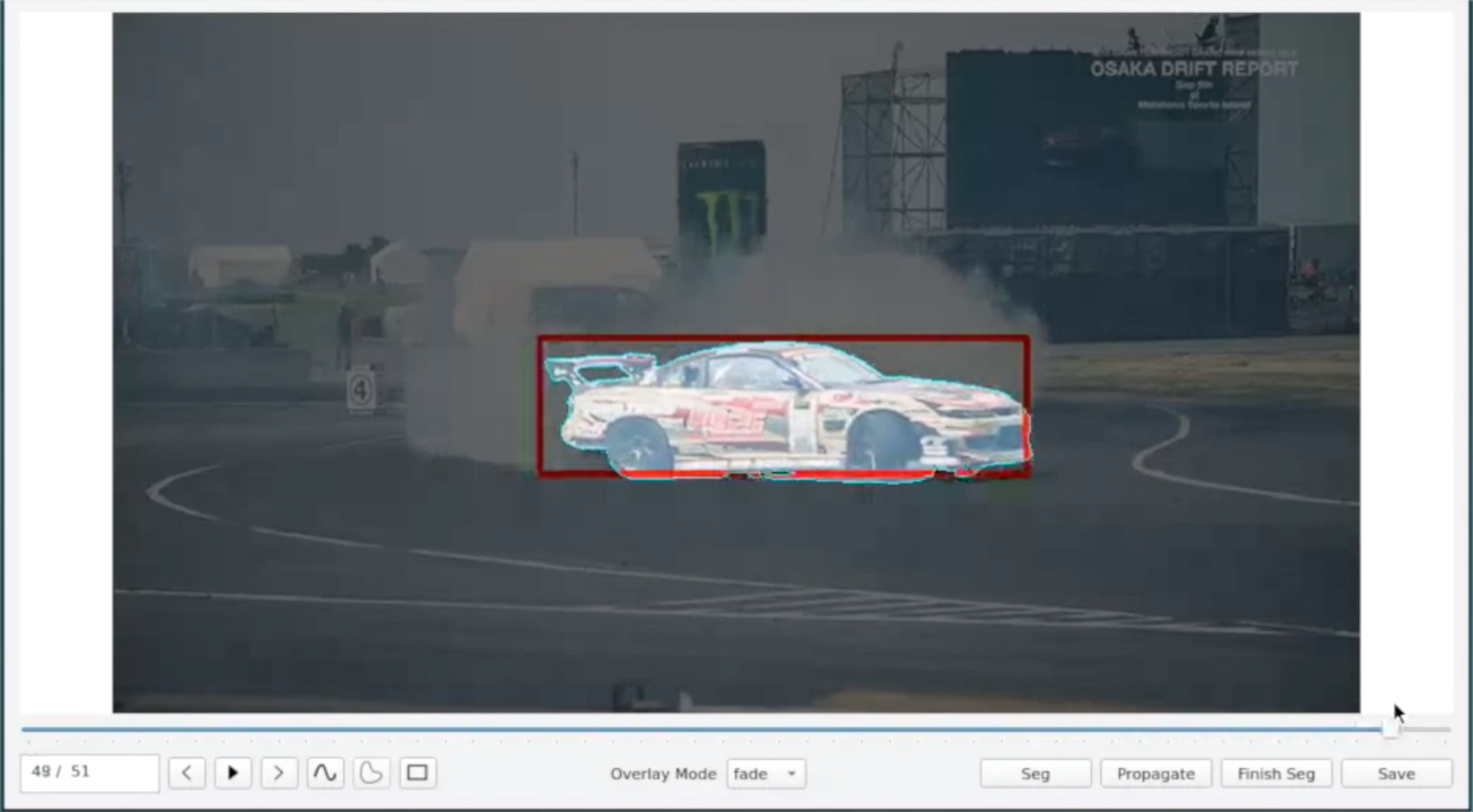}
\end{center}
\end{minipage}
\hspace{2mm}
\begin{minipage}{0.46\linewidth}
\vspace{3mm}
\textbf{Step5}: Done!
\end{minipage}

\caption{\footnotesize Step-by-step overview of Scribble-Box annotation tool. See video for seeing the tool being used in practice.}
\label{fig:tool}
\end{figure}

We demonstrate the step-by-step usage of our annotation tool in Fig~\ref{fig:tool}. Please refer to demo video from 00:47s.



\begin{figure*}[t!]
\begin{tabular}{ccc}
\includegraphics[width=0.33\linewidth, trim=0 0 0 0,clip]{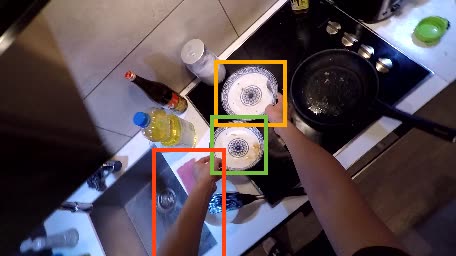} &
\includegraphics[width=0.33\linewidth, trim=0 0 0 0,clip]{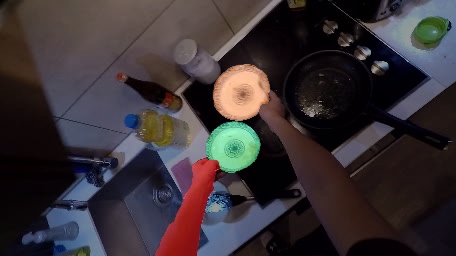} &
\includegraphics[width=0.33\linewidth, trim=0 0 0 0,clip]{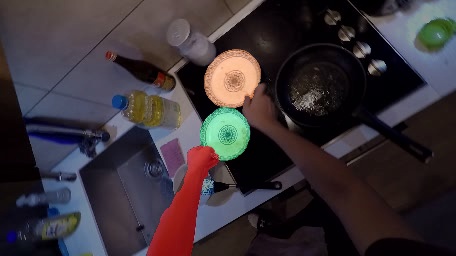} \\
\includegraphics[width=0.33\linewidth, trim=0 0 0 0,clip]{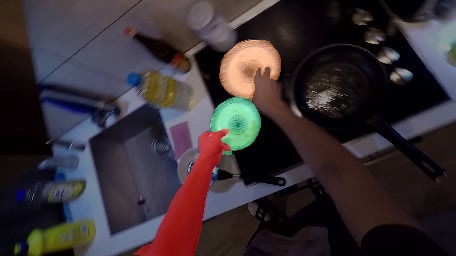} &
\includegraphics[width=0.33\linewidth, trim=0 0 0 0,clip]{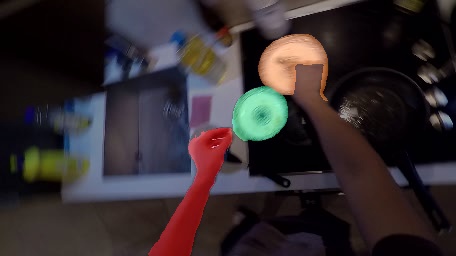} &
\includegraphics[width=0.33\linewidth, trim=0 0 0 0,clip]{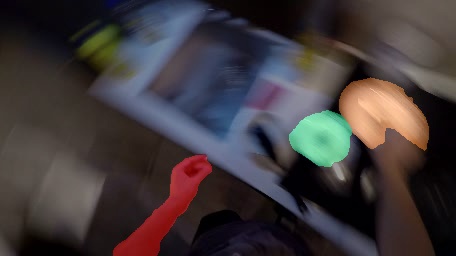} \\[-0.5mm]
\includegraphics[width=0.33\linewidth, trim=0 0 0 0,clip]{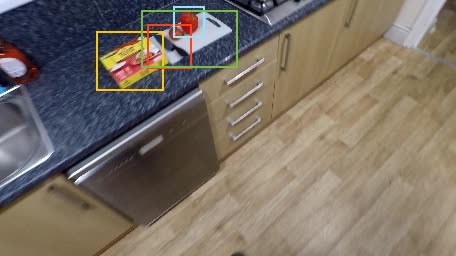} &
\includegraphics[width=0.33\linewidth, trim=0 0 0 0,clip]{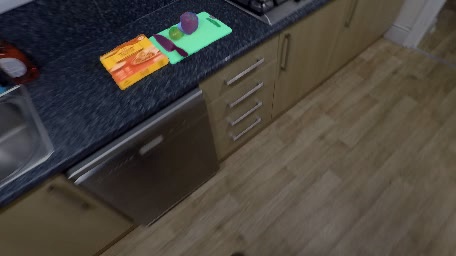} &
\includegraphics[width=0.33\linewidth, trim=0 0 0 0,clip]{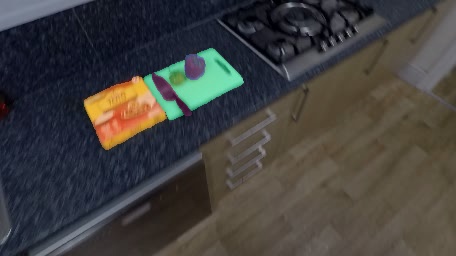} \\
\includegraphics[width=0.33\linewidth, trim=0 0 0 0,clip]{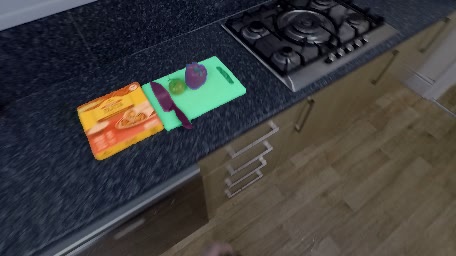} &
\includegraphics[width=0.33\linewidth, trim=0 0 0 0,clip]{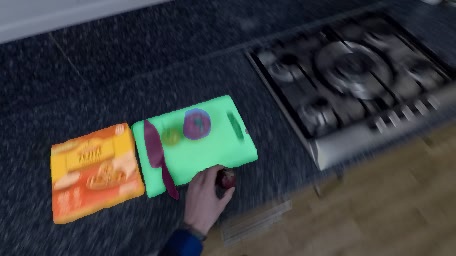} &
\includegraphics[width=0.33\linewidth, trim=0 0 0 0,clip]{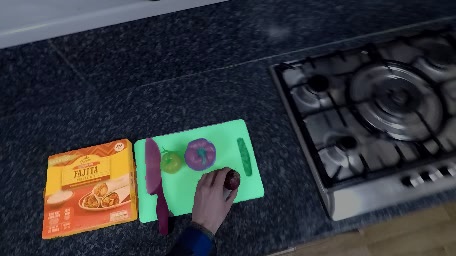} \\[-0.5mm]
\includegraphics[width=0.33\linewidth, trim=0 0 0 0,clip]{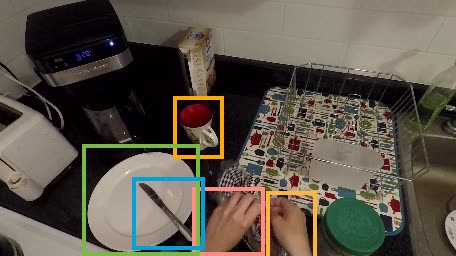} &
\includegraphics[width=0.33\linewidth, trim=0 0 0 0,clip]{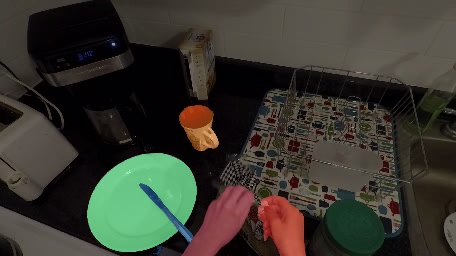} &
\includegraphics[width=0.33\linewidth, trim=0 0 0 0,clip]{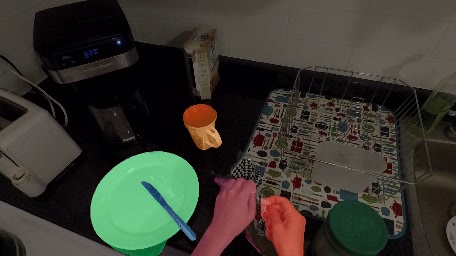} \\
\includegraphics[width=0.33\linewidth, trim=0 0 0 0,clip]{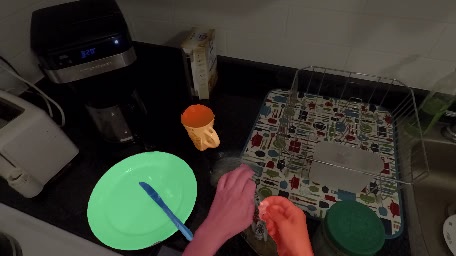} &
\includegraphics[width=0.33\linewidth, trim=0 0 0 0,clip]{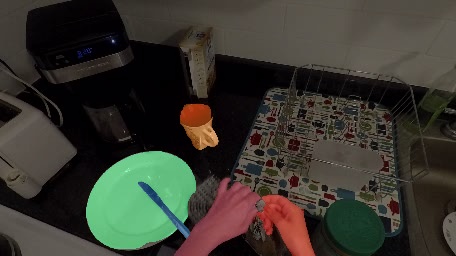} &
\includegraphics[width=0.33\linewidth, trim=0 0 0 0,clip]{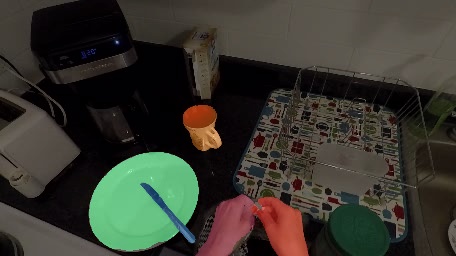} \\[-0.5mm]
\includegraphics[width=0.33\linewidth, trim=0 0 0 0,clip]{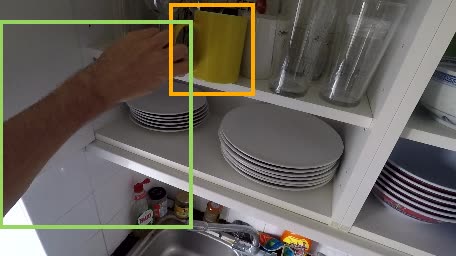} &
\includegraphics[width=0.33\linewidth, trim=0 0 0 0,clip]{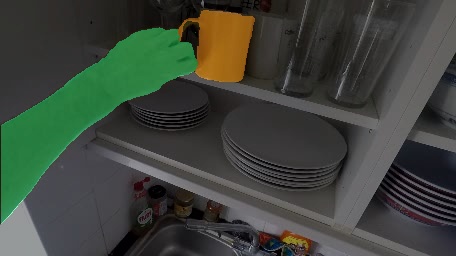} &
\includegraphics[width=0.33\linewidth, trim=0 0 0 0,clip]{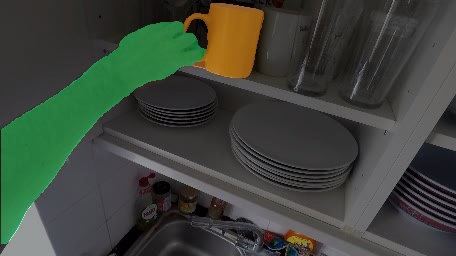} \\
\includegraphics[width=0.33\linewidth, trim=0 0 0 0,clip]{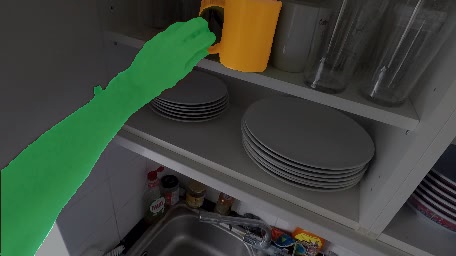} &
\includegraphics[width=0.33\linewidth, trim=0 0 0 0,clip]{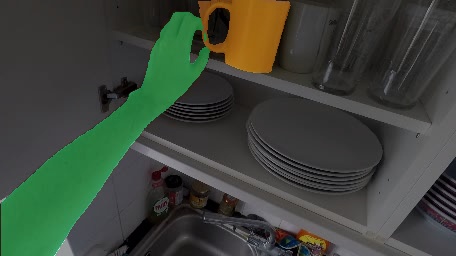} &
\includegraphics[width=0.33\linewidth, trim=0 0 0 0,clip]{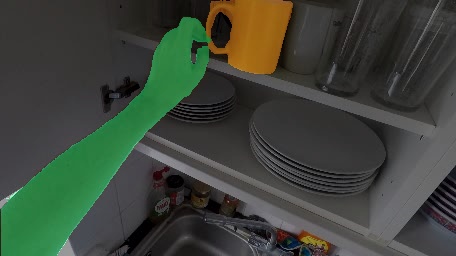} \\[-0.5mm]
\end{tabular}
\vspace{-4mm}
\caption{\footnotesize Example annotations using our annotation tool in practice on the EPIC-Kitchen dataset. Each object in a 100-frame video requiring on average 67.1s of annotation time (including inference time). The first column in each two rows indicates target objects.}
\label{fig:epic}
\vspace{-3mm}
\end{figure*}

\clearpage

\end{document}